\documentclass[10pt,twocolumn,letterpaper]{article}

\usepackage[pagenumbers]{cvpr} %

\definecolor{cvprblue}{rgb}{0.21,0.49,0.74}
\usepackage[pagebackref,breaklinks,colorlinks,allcolors=cvprblue]{hyperref}

\usepackage{array}
\usepackage{amsmath,amssymb,amsfonts}
\usepackage{algorithmic}
\usepackage{graphicx}
\usepackage{textcomp}
\usepackage{mathtools}
\usepackage{booktabs}
\usepackage{bbm}
\usepackage{wrapfig}
\usepackage{multirow}
\usepackage{comment}
\usepackage{tabularx}
\usepackage{color}
\usepackage{tikz}
\usepackage{colortbl}
\usepackage{caption}
\usetikzlibrary{calc}
\usetikzlibrary{positioning}

\renewcommand{\arraystretch}{1.3}

\newcommand{\parag}[1]{\noindent {\bf #1}}

\newcommand{\acro}{{\it PrIntMesh}}

\newif\ifdraft
\draftfalse

\ifdraft
  \newcommand{\PF}[1]{{\color{red}{\bf PF: #1}}}
  \newcommand{\pf}[1]{{\color{red} #1}}
  \newcommand{\DM}[1]{{\color{blue}{\bf JC: #1}}} 
  \newcommand{\dm}[1]{{\color{blue} #1}}
  \newcommand{\HL}[1]{{\color{orange}{\bf HL: #1}}} 

  \newcommand{\YH}[1]{{\color{dgreen}{\bf YH: #1}}}
  
\else
  \newcommand{\PF}[1]{}
  \newcommand{\pf}[1]{#1}
  \newcommand{\HL}[1]{}
  
  \newcommand{\YH}[1]{}
  
  \newcommand{\DM}[1]{}
  \newcommand{\dm}[1]{#1}
\fi

\title{PrIntMesh: Precise Intersection Surfaces for 3D Organ Mesh Reconstruction}

\author{Deniz Sayin Mercadier\smash{\textsuperscript{1}},%
Hieu Le\smash{\textsuperscript{2}},%
Yihong Chen\smash{\textsuperscript{1}},%
Jiancheng Yang\smash{\textsuperscript{3}},%
Udaranga Wickramasinghe\smash{\textsuperscript{4}},%
Pascal Fua\smash{\textsuperscript{1}}\\
\smash{\textsuperscript{1}}EPFL \quad
\smash{\textsuperscript{2}}UNC Charlotte \quad
\smash{\textsuperscript{3}}ELLIS Institute Finland \quad
\smash{\textsuperscript{4}}Adis SA, Lausanne, Switzerland\\
}

\begin{document}
\maketitle

\begin{abstract}

Human organs are composed of interconnected substructures whose geometry and spatial relationships constrain one another. Yet, most deep-learning approaches treat these parts independently, producing anatomically implausible reconstructions. We introduce \acro{}, a template-based, topology-preserving framework that reconstructs organs as unified systems. Starting from a connected template, \acro{} jointly deforms all substructures to match patient-specific anatomy, while explicitly preserving internal boundaries and enforcing smooth, artifact-free surfaces. We demonstrate its effectiveness on the heart, hippocampus, and lungs, achieving high geometric accuracy, correct topology, and robust performance even with limited or noisy training data. Compared to voxel- and surface-based methods, \acro{} better reconstructs shared interfaces, maintains structural consistency, and provides a data-efficient solution suitable for clinical use.

\end{abstract}

\section{Introduction}

\begin{figure}[th!]
    \centering
    \begin{subfigure}{0.5\textwidth}
        \centering
        \includegraphics[width=\textwidth]{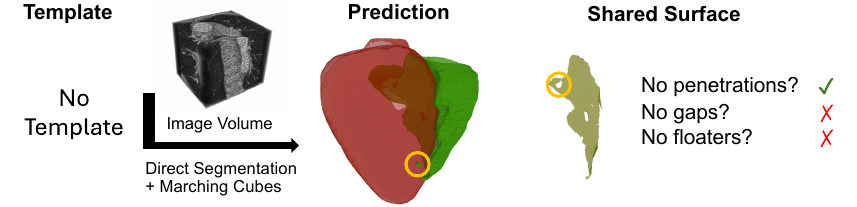}
        \caption{Voxel-based approach}
        \label{fig:fig1-voxel}
    \end{subfigure}\\%
    \begin{subfigure}{0.5\textwidth}
        \centering
        \includegraphics[width=\textwidth]{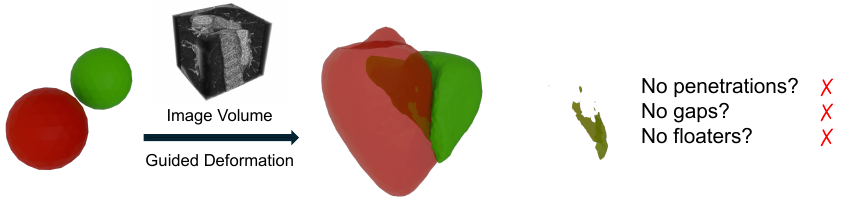}
        \caption{Independent mesh-based approach}
        \label{fig:fig1-mesh}
    \end{subfigure}\\%
    \begin{subfigure}{0.5\textwidth}
        \centering
        \includegraphics[width=\textwidth]{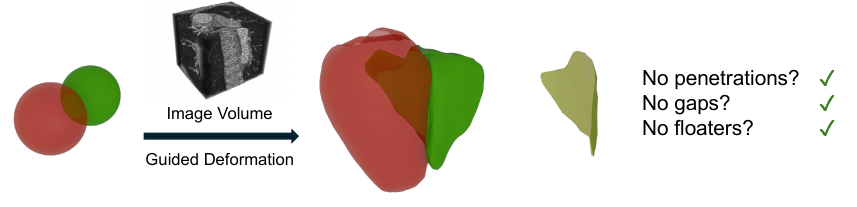}
        \caption{Our approach}
        \label{fig:fig1-ours}
    \end{subfigure}
    \vspace{-6mm}
    \caption{\textbf{Voxel-based vs Mesh-based approaches.}
    (a) Voxel-based approaches such as nnU-Net~\cite{Isensee21} currently are the dominant ones for organ segmentation. However, they can yield gaps, floaters, and jagged boundaries, requiring post-processing to produce usable meshes. 
    (b) Existing mesh-based methods generate smooth surfaces but model each class separately, leading to interpenetrations and misaligned boundaries between components. 
    (c) \acro{} deforms a joint template for all components and precisely reconstructs shared surfaces, producing a smooth, watertight and topologically correct mesh, without post-processing.}
    \label{fig:intro}
\end{figure}

Human organs are not single solids but interconnected systems. The heart, for example, consists of four chambers—2 atria and 2 ventricles—separated by thin walls, linked by valves, and threaded with vessels. \pf{These shapes are tightly coupled; the deformations of one impact the others}. Modeling the heart thus requires capturing these relationships, not just segmenting individual chambers. Even small misalignments can compromise anatomical plausibility.

Most current deep-learning approaches ignore this. They break organs apart, predicting substructures such as the left atrium or septal wall independently and ignoring their spatial or functional interdependence. This disrupts continuity and limits use in downstream tasks such as biomechanics~\cite{updegrove2017simvascular,salvador2024whole}, electrophysiology~\cite{li2024solving,sakata2025digital}, and surgical planning~\cite{sakata2024assessing}.
Voxel-based models like nnU-Net~\cite{Isensee21} assign labels independently across the volume, often leaving gaps and disconnected fragments, as in Fig.~\ref{fig:fig1-voxel}. Surface-based methods~\cite{Wickramasinghe20,Kong21a,Kong21b,Bongratz22,Yang23b} reconstruct each chamber separately, with no guarantee of proper alignment at shared interfaces, as in Fig.~\ref{fig:fig1-mesh}. The resulting meshes are often anatomically implausible and clinically unreliable.

To remedy this, we propose \acro{}, a method that reconstructs organs as unified, connected systems rather than collections of parts. Instead of attempting to reconcile independently predicted substructures as most earlier approaches do, \acro{} starts from a template that encodes full anatomical connectivity. It acts as a topological scaffold, ensuring structural integrity throughout deformation. Guided by volumetric image features, the network deforms this scaffold to match individual anatomy while preserving its internal organization.

\dm{\acro{} comprises two modules, a feature extractor that processes volumetric imagery and a mesh deformation network that deforms the whole connected template. Unlike earlier models that predict each component independently~\cite{Wickramasinghe20,Kong21a}, it deforms all components together, preserving spatial relationships by construction. Anatomical realism is maintained by explicitly supervising pairwise interfaces between substructures to preserve coherent internal walls, and by applying geometric regularization terms that encourage smooth, artifact-free surfaces.}

On high-resolution cardiac data, \acro{} achieves both high geometric accuracy and guaranteed topological consistency—a combination unmatched by competing methods. While voxel models may yield similar voxelwise scores, they fail to maintain connectivity or realistic internal boundaries.
We further tested \acro{} on hippocampus and lung datasets, demonstrating generality across anatomies with distinct topologies—from the folded hippocampus to disconnected lungs with imperfect annotations. In all cases, \acro{} produces structurally valid, smooth reconstructions even with limited training data.

Crucially, the structural prior embedded in the template enables data-efficient learning. \pf{For lungs, \acro{} yields a 2–3× lower Chamfer error than nnU-Net~\cite{Isensee21} while guaranteeing topological correctness and requiring fewer than 100 samples}—highlighting its practicality for clinical deployment, with scarce annotated data and rapid adaptation required.

\noindent In short, our contributions are as follows:
\begin{itemize}
 
  \item We propose a template-based, topology-preserving framework that reconstructs organs as unified, connected systems rather than independent parts.

  \item We \pf{use our templates to strictly} enforce anatomical interfaces and geometric regularity, ensuring smooth, artifact-free surfaces and correct internal boundaries.

  \item We demonstrate high-accuracy, data-efficient reconstruction across diverse anatomies—including heart, hippocampus, and lungs—even given limited or imperfect training data.

\end{itemize}
The code will be made publicly available.

\section{Related Work}

Deep learning has become the de facto standard for medical image segmentation, including segmenting the liver~\cite{Zhang23e}, the whole heart~\cite{Yang23b}, and brain tumors~\cite{Zeng23,She23}. Most methods output voxel-wise labels using volumetric networks such as nnU-Net~\cite{Isensee21} or atlas-based registration approaches~\cite{Bai13b,Yang18f,Iglesias15}. These representations assign a single label per voxel, making it impossible to directly model shared surfaces or multi-part structures. While this may not severely affect standard metrics like Dice, it often leads to topological errors requiring post-processing.

Topology, however, is crucial for downstream modeling. Hemodynamic simulations demand watertight, connected geometries for stable CFD~\cite{updegrove2017simvascular}; biomechanics modeling relies on anatomically continuous walls to predict stresses~\cite{salvador2024whole}; electrophysiological models require consistent atrial and ventricular meshes for accurate excitation propagation~\cite{li2024solving,sakata2025digital}; and surgical planning or disease progression studies depend on coherent 3D anatomy~\cite{sakata2024assessing,pak2024robust}. As large-scale digital twins emerge~\cite{qian2025developing,qiao2025personalized}, manual mesh repair usually involving  voxel-to-surface conversion followed by manual correction~\cite{Fischl12,Charton21}, becomes infeasible.

Implicit representations based on Signed Distance Fields (SDFs)~\cite{Yang24a,Verhulsdonk24,Le23a} offer an alternative by learning latent shape priors refined to match voxel segmentations. However, these methods require topologically correct training data or geometric losses, lack hard topology guarantees, and remain computationally expensive since the refinement occurs at inference time.

Another direction is to directly predict triangulated meshes~\cite{Wickramasinghe20,Bongratz22,Kong21b,Yang23b} when internal structures are irrelevant, which yields accurate boundaries. When they are important as in the heart, separate templates are often used per chamber~\cite{Wickramasinghe20,Kong21a,Kong21b,Bongratz22,Yang23b}, sometimes with per-template scaling or translation. Independent deformation of these templates, however, causes interpenetrations and gaps that are difficult to fix post hoc.

In contrast, our approach deforms a single unified template encoding internal structures, enabling topologically consistent multi-component reconstructions in a single forward pass without post-processing.

\section{Method}

Our method reconstructs anatomically consistent multi-part organ surfaces from volumetric medical images using a single, unified mesh template. This template encodes the correct topology of the organ and is deformed using learned image features to fit a specific patient's anatomy.
The  key components of our approach are:
\begin{enumerate}
    \item A \textbf{topologically-correct mesh template} that models all substructures and their interfaces jointly.
    \item A \textbf{feature-guided mesh deformation network} that aligns and adapts this template to medical imaging data.
    \item A \textbf{training strategy with combined loss functions} that encourages both overall shape accuracy and precise inter-substructure surface modeling.
\end{enumerate}
We describe each one below, using heart reconstruction for illustration purposes. The method applies similarly hippocampus and lung reconstruction by simply changing the template we deform. 

\subsection{Building a Topologically Correct Template}
\label{sec:template}

Our algorithm starts with a template that captures the topological structure of the target organ. For the heart, this includes the four chambers—left/right ventricles and atria—connected by biologically plausible shared walls, such as the myocardium and septa. Our approach to designing the template is  to start from a basic primitive geometry that naturally supports defining interconnected components. 

To this end, we use a rhombicuboctahedron~\cite{Kepler19}, illustrated in Fig.~\ref{fig:template1}(a). It has 26 faces, consisting of 8 equilateral triangles and 18 squares, which makes it a good approximation of a sphere while also being easy to connect to other rhombicuboctahedra by having them share their square faces. We exploit this property by positioning four rhombicuboctahedra that share faces as shown in Fig.~\ref{fig:template1}(b). We then subdivide all faces into smaller facets labeled with a class label denoting one or more substructures. Finally we adjust the position of individual vertices to make the four chambers initially spherical. This yields the initial template of Fig.~\ref{fig:template1}(c), which  reflects the anatomical layout, that is, adjoining chambers, within a single, connected mesh. 

{
\newcommand{\arrw}{0.40cm}
\newcommand{\tempwidth}{0.14\textwidth}

\begin{figure}[th]
    \centering
    \begin{tikzpicture}
        \node[inner sep=0] (a) {
            \subfloat[][ \label{fig:rhomb}]{
                \includegraphics[width=\tempwidth]{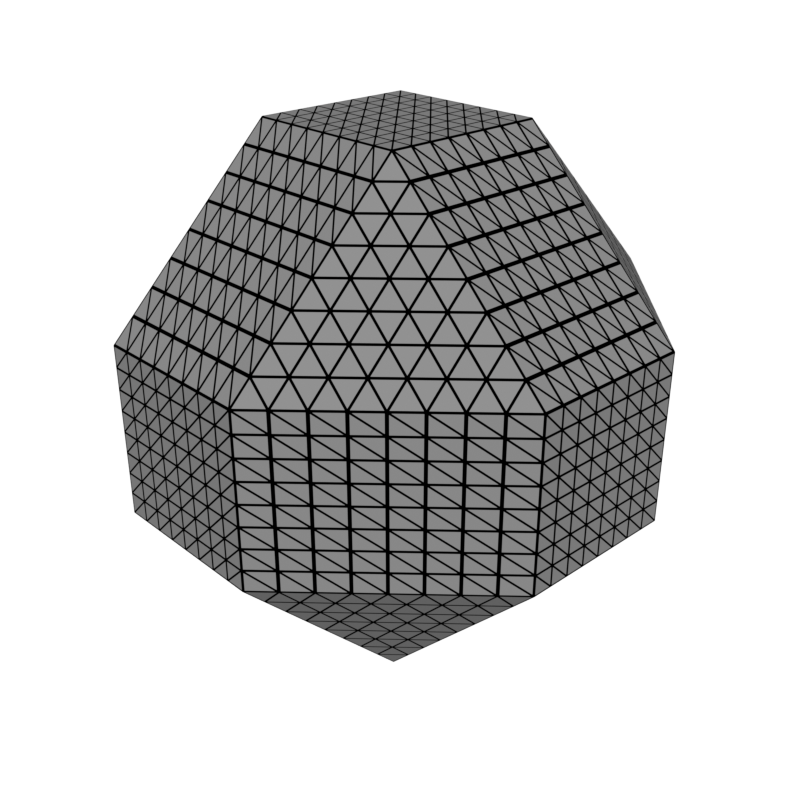}
            }
        };

        \node[inner sep=0,right=\arrw of a] (b) {
            \subfloat[][ \label{fig:rhomb4}]{
                \includegraphics[width=\tempwidth]{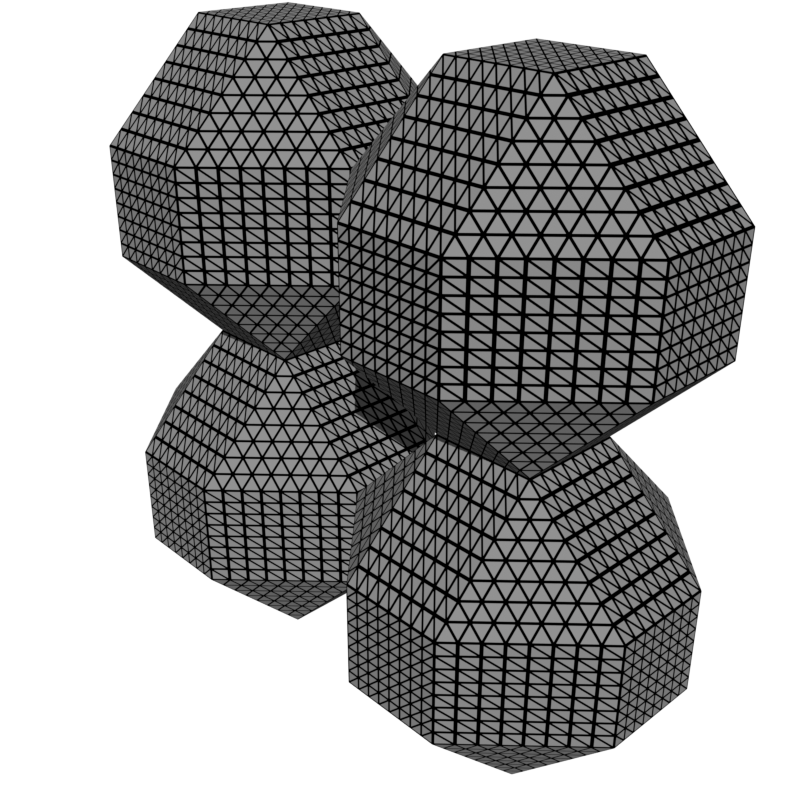}
            }
        };

        \node[inner sep=0,right=\arrw of b] (c) {
            \subfloat[][ \label{fig:sphere4}]{
                \includegraphics[width=\tempwidth]{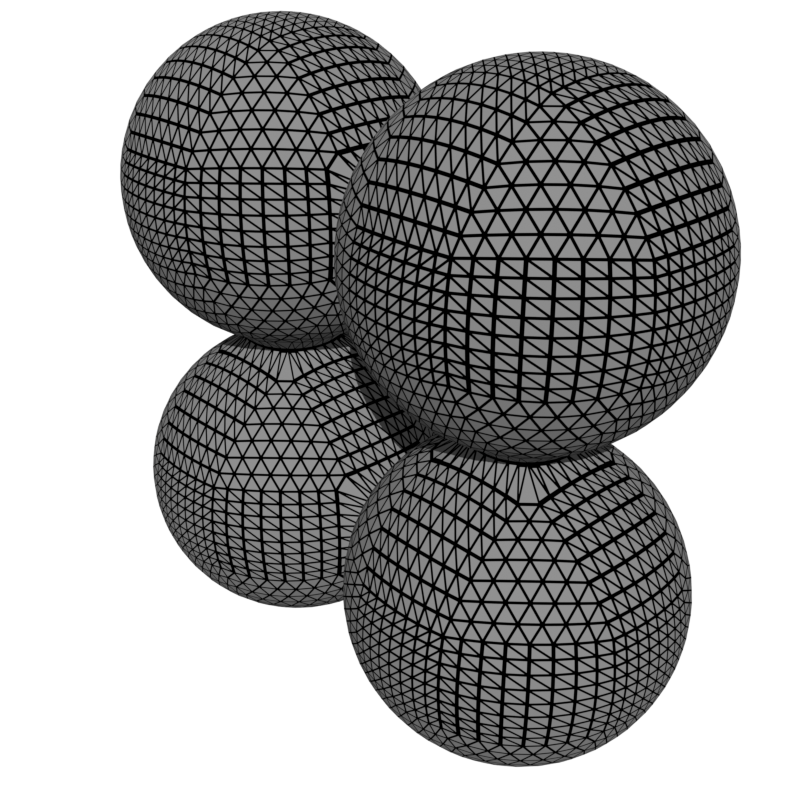}
            }
        };

        \draw[->,thick] (a.east) -- (b.west);
        \draw[->,thick] (b.east) -- (c.west);
    \end{tikzpicture}

    \caption{\protect\textbf{Building the heart template.} 
    (a) Initial rhombicuboctahedron. 
    (b) Four rhombicuboctahedra glued together. 
    (c) Their vertices are uniformly displaced to create four spheres.}
    \label{fig:template1}
\end{figure}

For the hippocampus and its two parts, we built the template of Fig.~\ref{fig:other-templates}(a), using only two spheres sharing one intersection instead of four. For the lung template of Fig.~\ref{fig:other-templates}(b), we used two spheres for the two lobes of the right lung and three connected ones for the three lobes of the left lung.

\begin{figure}[th]
    \vspace{-7mm}
    \centering
    \subfloat[][\label{fig:hippo-template}]{
        \includegraphics[width=0.21\textwidth]{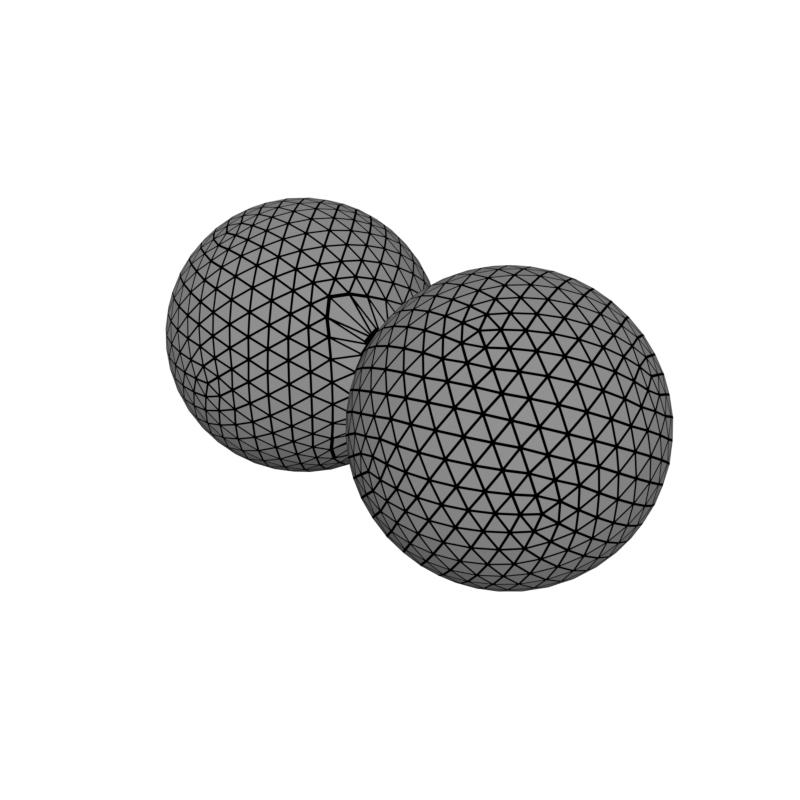}\vspace{-3mm}
    }
    \subfloat[][\label{fig:lung-template}]{
        \includegraphics[width=0.21\textwidth]{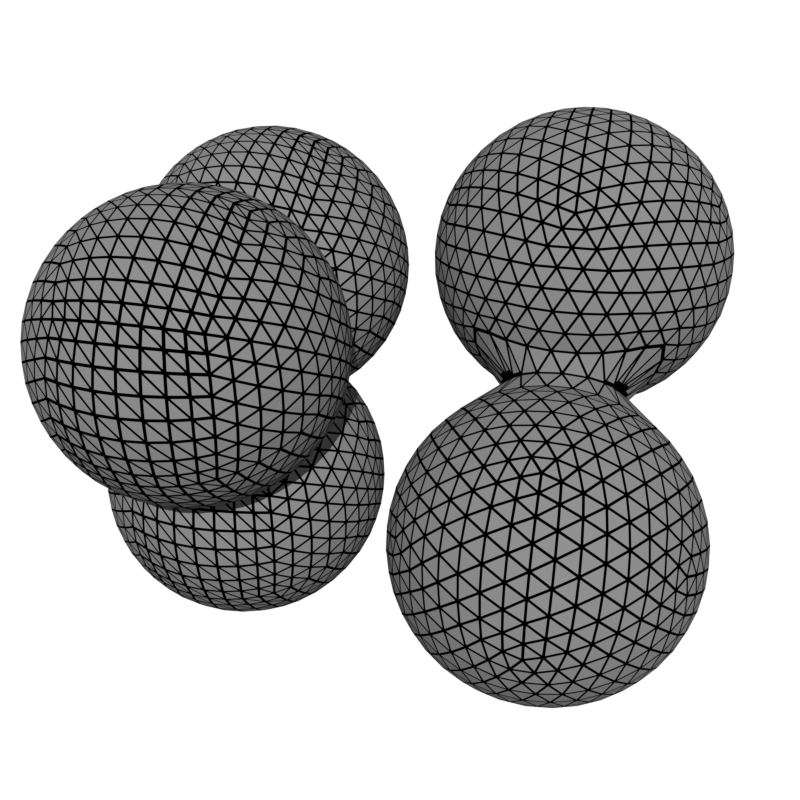}\vspace{-3mm}
    }
     \vspace{-1mm}
    \caption{Templates for (a) the hippocampus and (b) the lungs.}
    \label{fig:other-templates}
\end{figure}

\begin{figure*}[th]
    \centering
    \includegraphics[width=\linewidth]{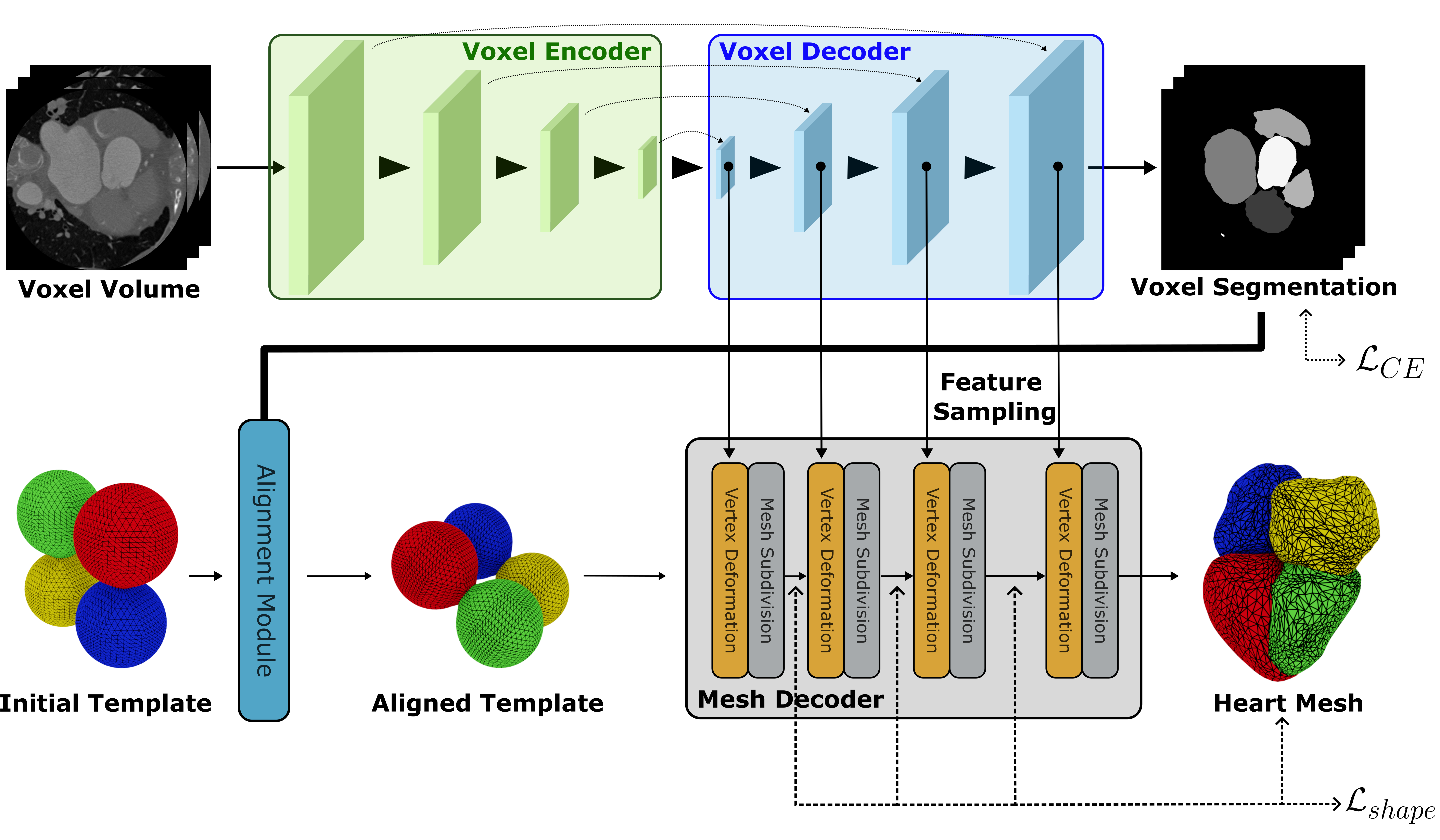}
      \vspace{-8mm}
    \caption{\protect\textbf{Network Architecture. }The voxel encoder and decoder follow a U-Net architecture including 
    skip-connections, and are trained with cross-entropy loss. The initial heart template goes through a simple alignment
    procedure based on the segmentation, and is then deformed and subdivided in multiple steps by the mesh decoder. The mesh
    decoder samples features from the voxel decoder to apply correct deformations, and the shape loss is applied to each 
    intermediate mesh.}
    \label{fig:arch}
\end{figure*}
 
\vspace{-2mm}
\subsection{Deforming the Template to Match the Data}

A correct, unified template is our starting point. A core challenge then lies in deforming it to accurately reconstruct the anatomy of a specific patient from volumetric medical images, while preserving both topology and mesh regularity as described below. To this end, we use the two-stream network architecture of  Fig.~\ref{fig:arch} to compute translation vectors for each template vertex by combining voxel-based semantic understanding with mesh-based surface generation, as in Voxel2Mesh~\cite{Wickramasinghe20}. 

\subsubsection{Network Architecture}

The first stream is a 3D U-Net~\cite{Ronneberger15} that outputs a dense voxel-wise segmentation and multi-scale feature maps. These maps are fed to the second stream to guide the mesh deformation and to compute an affine transformation that coarsely aligns the template to the target anatomy. Because our template is not radially symmetric, this alignment is critical to avoid unnecessary deformation. We estimate the transform from the predicted class centroids of the segmentation; the computation is fast, differentiable, and performed at every training iteration. 

The second stream is a mesh decoder that deforms the aligned template in a coarse-to-fine fashion. At each stage, features are sampled from the voxel stream and used to update vertex positions. Instead of learning curvature-aware unpooling as in~\cite{Wickramasinghe20}, we adopt a simpler subdivision scheme with specific layers quadrupling the number of triangles by splitting each face uniformly. This allows the mesh to gradually increase resolution while preserving class labels on vertices and faces, including those on shared interfaces.

To implement both streams, we started from the publicly available Voxel2Mesh code~\cite{Wickramasinghe20} and modernized it by replacing the hand-designed adjacency matrix based graph convolutions by newer, more efficient {\it PyTorch 3D} versions~\cite{Ravi20}. This greatly reduced memory usage and running time, allowing us to work with larger input image resolutions and higher output vertex counts.

\subsubsection{Preserving Topology}
\label{sec:topology}

Unconstrained deformation would yield anatomically invalid results, especially for shared surface such as the septal walls between heart chambers. This is because each component is encouraged to match its own target shape based on local image evidence. Without constraints enforcing agreement across interfaces, these shared surfaces can be pulled in conflicting directions, leading to drift, misalignment, or collapse. To prevent this, we  explicitly model and supervise the surfaces shared  between anatomical structures, such as the wall between the left and right ventricles, and not just the individual structures. Each shared surface is represented as a distinct region in the unified template mesh and is assigned a dedicated label. This shifts the learning objective: The network is not simply encouraged to place adjacent structures near each other, but to reconstruct their exact shared interface. The result is a single, continuous surface that inherently preserves the topology of both the individual parts and their points of contact.

These regions are supervised directly during training, just like the primary anatomical components by minimizing a Chamfer-based distance shape matching loss 
 \begin{align}
     \mathcal{L}^l_{\text{match}}  = \sum_{c=1}^C \mathcal{L}^{l,c}_{\text{Chamfer}} \; , 
 \end{align}
 where the index $l$ denotes the fact that the losses are evaluated after each intermediate stage of the shape decode and averaged, instead of only being computed on the final mesh.

Minimizing the Chamfer term drives geometric alignment between the predicted mesh and the ground truth. Since the mesh represents all substructures and their interfaces, we compute the Chamfer loss separately for each labeled region $c$, including both anatomical parts and intersection surfaces. At training time, we randomly sample points from both predicted and ground truth meshes and compute a symmetric Chamfer distance per region. This helps the model learn accurate geometry for both individual structures and shared surfaces. Each class is weighted equally to assign sufficient importance to the shared surfaces, whose contribution would otherwise be dwarfed due to being significantly smaller than the primary components.

\subsubsection{Preserving Mesh Regularity}
\label{sec:geometry}

In addition to preserving topology, geometric integrity must also be maintained. Without it, one is likely to observe vertex collapse, irregular triangle areas, and tangled regions, especially near points of contact between substructures. This would not only degrade reconstruction quality but also render the mesh unsuitable for downstream tasks such as simulation or quantitative analysis. Critically, once such degradations occur, they are difficult or impossible to repair in post-processing. To eliminate such artifacts, we introduce a geometric regularization loss $ \mathcal{L}_{\text{reg}}$ that, when minimized, forces the preservation of mesh quality by ensuring overall smoothness, avoiding vertex collapse, and promoting a uniform triangle structure. This eliminates the need for post hoc alignment or mesh repair steps, which are typically error-prone and inconsistent. We take this loss to be
 \begin{align}
 \mathcal{L}_{\text{reg}}^l  &= \lambda_1 \mathcal{L}^l_{\text{Edge}} + \lambda_2  \mathcal{L}_{\text{EdgeUnif}}^l  + \lambda_4 \mathcal{L}_{\text{Norm}}^l + \lambda_5   \mathcal{L}_{\text{Lapl}}^l \; ,
 \end{align}
 where $l$ again denotes the stage of the mesh decoder, the  $\lambda$ coefficients are scalar weights, and $\mathcal{L}^l_{\text{Edge}}$,  , $\mathcal{L}_{\text{Norm}}^l$, and  $\mathcal{L}_{\text{Lapl}}^l$ \dm{are standard regularization terms that we define in the supplementary.  $\mathcal{L}_{\text{EdgeUnif}}^l$ is specific to our approach and penalizes edge lengths deviations to keep the triangles uniform.}

\begin{align}
     \mathcal{L}_{\text{EdgeUnif}}^l & = \sqrt{ \frac{1}{E-1} \sum_{i=1}^E (|e_i| - \mathcal{L}^l_{\text{Edge}})^2 } \; ,  \nonumber
\end{align}

Because our component submeshes are open surfaces and do not lend themselves well to smoothing by regularizers designed for closed surfaces, we compute these losses on the full mesh, rather than for each class separately, as we do for $ \mathcal{L}^{l,c}_{Chamfer}$ in Section~\ref{sec:topology}.  Shared vertices receive contributions from neighbors in different sub-meshes, with appropriate weights. 

\subsubsection{Global Loss}

In the previous sections we introduced $\mathcal{L}_{\text{reg}}^l$ and $\mathcal{L}^l_{\text{match}}$ whose minimization promotes mesh regularity and topology preservation. To encourage an accurate fit to the training annotations, we also define $\mathcal{L}_{CE}$, the cross-entropy loss, and $\mathcal{L}_{Dice}$ is the Dice loss ~\cite{Falk18} with respect to the ground truth, both computed voxel-wise. When training the network, we minimize the global loss 
\begin{align}
    \mathcal{L} &=  \! \mathcal{L}_{CE}   \!  +  \! \mathcal{L}_{Dice}  \!  +  \! \frac{1}{L} \sum_{l=1}^L \left ( \lambda_{\text{match}} \mathcal{L}^l_{\text{match}} + \lambda_{\text{reg}} \mathcal{L}^l_{\text{reg}} \right ) . 
\end{align}

\section{Experiments}

We use three datasets to validate our approach on heart, lung, and hippocampus reconstruction, all of which incorporate internal structures that need to be modeled accurately. The template we use for heart reconstruction has been introduced in Section~\ref{sec:template} and is depicted by Fig.~\ref{fig:template1}. The hippocampus and lung templates are shown  in Fig.~\ref{fig:other-templates}. \dm{Further preprocessing and architecture details are provided in the supplementary. }

\subsection{Datasets}
 
\noindent {\bf MM-WHS Heart Dataset}~\cite{Zhuang18a}. 
It contains 20 whole-heart annotated CT scans for training and 40 CT scans, without  publicly available annotations. Seven classes are defined: the left ventricle blood cavity (LV-),  the right ventricle blood cavity (RV), the left atrium blood cavity (LA), the right atrium blood cavity (RA), the myocardium of the left ventricle (Myo), the ascending aorta (Ao), and the pulmonary artery (PA). In our experiments, we only model the four heart chambers. Thus,  we  combine left ventricle blood cavity (LV-) and its  myocardium (Myo) into a single left ventricle class (LV), and ignore the Ao and PA classes. We also define three {\it intersection classes} that correspond to the surfaces shared  by  adjacent substructures. Specifically we consider LV $\cap$ RV, LV $\cap$ LA, and RV $\cap$ RA but not  LA $\cap$ RA because the atrial septum is not consistently labeled in the dataset due to the annotations being intended for blood cavities rather than the complete atrial structure. This yields a total of seven classes that we explicitly model.

Without test set annotations, we cannot directly compute metrics on it. Instead, we perform 5-fold cross-validation on the training set, splitting into 5 sets of 16 training and 4 validation samples. We train separate models for each set and report the mean and standard deviation for each metric. 

\parag{MSD Hippocampus Dataset}~\cite{Antonelli2022}.  
It comes from the Medical Segmentation Decathlon and comprises 260 small voxel volumes from MRI scans, with the hippocampus split into its anterior and posterior components. We define Anterior $\cap$ Posterior as the only intersection class, which yields a total of three distinct classes. We  split the dataset into  200 training samples and 60 test ones.

\parag{TS Lung Dataset}~\cite{Wasserthal23}. 
For lung reconstruction, we use images and annotations from the  TotalSegmentator data. It comprises 1228 CT scans annotated and refined using an active learning setup with a human-in-the-loop. We extract those having viable lung annotations, yielding 348 annotated lung scans  Five classes are defined, the lower, middle and upper right lobe, which we designate as LR, MR and UR, and the lower and upper left lobe, designated as LL and UL. We also consider the four intersection classes LR $\cap$ MR, LR $\cap$ UR, MR $\cap$ UR and LL $\cap$ UL, yielding nine separate classes.

\subsection{Baselines}

We compare against three state-of-the-art baselines that epitomize the three main approaches used to produce 3D segmentations, that is, relying on volumetric segmentation, triangulated surface meshes, or tetrahedral meshes. 

\parag{nnU-Net}~\cite{Isensee21}.  It is a state-of-the-art voxel segmentation model, recently updated with deeper residual models~\cite{Isensee24}. As such, it cannot generate self-intersections, but is still subject to inconsistencies such as "floaters", that is, connected components separated from the main model, leading to potential gaps and holes in intersecting regions.

\parag{MeshDeformNet}~\cite{Kong21a}. It is a multi-class improvement over the original Voxel2Mesh~\cite{Wickramasinghe20} that focuses on the heart. Instead of using the same spherical template for each class, it uses a per-class translation and scaling factor in advance, which makes the template slightly different for each class. As a mesh-based method, it can generate erroneous intersections between classes, but not nnU-Net-like floaters. 

\parag{MedTet}~\cite{Chen24f}. It is a recent method that relies on deep-marching tetrahedra~\cite{Shen21a} instead of 3D meshes to model organs. This tends to yield better  reconstruction accuracy 
thanks to an hybrid vertex grid and signed-distance approach. However, as the outputs are still independent tetrahedral meshes, it can still make the same topological errors as the mesh-based approaches.

\subsection{Evaluation Metrics}

The primary goal of our approach is to produce anatomically faithful 3D reconstructions with correct topology—maintaining the intended connectivity between structures, avoiding interpenetration, and preserving shared surfaces. These structural properties are essential for both clinical relevance and downstream applications such as simulation. However, they are not adequately captured by standard segmentation metrics, which typically focus on voxel-wise or surface-level similarity. To fully assess reconstruction quality, we report both \textbf{topology-aware metrics}, which directly evaluate structural correctness, and \textbf{standard metrics}, which provide complementary assessments of overall volume alignment. This dual evaluation highlights the limitations of conventional metrics in avoiding anatomical errors and underscores the strengths of our method in preserving topological integrity. We describe both kinds below.

\subsubsection{Topology Metrics} 

Topological correctness is difficult to assess using standard metrics. Structural errors such as unintended overlaps or missing connections often have little effect on global scores like Dice or Chamfer, which average over entire volumes or surfaces. Localized shifts at boundaries may only impact a small number of voxels, but these small inconsistencies can lead to artifacts that compromise anatomical validity and limit the utility of the reconstruction in downstream tasks.

Such topological errors arise in different ways depending on the representation. Mesh-based methods may generate interpenetrating surfaces between structures that should only be adjacent, while voxel-based methods—particularly those that segment each structure independently—can leave small gaps or disconnected regions where continuity is required. \pf{ Because enforcing topological correctness has not been a primary focus in the field of organ reconstruction, there are no standard metrics to quantify topological mistakes. Thus, we designed two such  metrics to quantify structural inconsistencies at interfaces:} one measuring the volume of mesh intersections, and another capturing the presence of undesired gaps between components.

\parag{Volume Intersections.} For mesh-based methods that generate erroneous intersections, we use mesh boolean operations to extract the intersection between meshes corresponding to different classes and compute its volume relative to the total mesh volume. 

\parag{Unwanted Gaps.} Voxel-based methods can produce unintended empty regions inside the heart due to disconnected or misaligned surfaces. We quantify these artifacts by counting the number of isolated background (empty) connected components that are entirely enclosed within the structure—excluding the outer background. In this context, each such enclosed component corresponds to a "hole" or gap. 

\subsubsection{Standard Metrics} In addition to topology metrics, we report standard evaluation metrics commonly used in segmentation benchmarks. While these do not directly assess structural integrity, they offer a baseline for geometric similarity. They come in two main flavors, volumetric and surface-based. Thus we consider the following two. 

\parag{Chamfer Distance.}  It is calculated between two point clouds. For the mesh-based methods,  we obtain them by sampling 1024K points from the output surfaces. For the volumetric ground-truth data and the output of nnU-Net, we get them by running the Marching Cube algorithm~\cite{Lorensen87}.  We calculate it for every class, including interfaces between substructures. With intersection surfaces not explicitly modeled by either MeshDeformNet or MedTet, we use boolean operations to extract approximate ones from the meshes to compute Chamfer distances. All calculations are done with vertex coordinates normalized into the [-1, 1] range, and we report values multiplied by $10^{-3}$. %

\parag{Normal Consistency}~\cite{Gkioxari19}. It is calculated in a very similar way to Chamfer distance by comparing the orientation of surface normals at matching points rather than their distance. Based on the cosine distance, its best value is 1 when the normals match perfectly, and 0 when they are completely orthogonal. Normals at random points on the mesh surface are obtained by interpolating vertex normals. 

\parag{Dice Coefficient}~\cite{Dice45}. It measures volumetric overlap between ground truth and prediction, while being more sensitive to class imbalance than the Jaccard index~\cite{Bertels19a}. 
We compute it for each segmentation class. For mesh based methods such as ours, we first convert the mesh representation into a voxel-based one to compute this metric.

\subsection{Comparative Results}

\newcolumntype{?}{!{\vrule width 1pt}}

\begingroup
\begin{table*}
    \caption{Per-class Chamfer distance ($\times 10^{-3})$ and normal consistency on MM-WHS-4. \textit{Blue denotes the best method and red the second best. Our method is consistently one of the two and, when it is second-best, it is only by a small margin.}}
    \vspace{-3mm}
\label{tab:surface-heart}
\centering
\renewcommand{\arraystretch}{1.2}
\resizebox{\textwidth}{!}{
\begin{tabular}{@{}lcccccccccc}
\hline

& \multicolumn{8}{c}{\textbf{Intersection Classes}} & \multicolumn{2}{c}{\textbf{Other Classes}} \\
\cmidrule(lr){2-9} \cmidrule(l){10-11}

& \multicolumn{2}{c}{LV $\cap$ LA} 
    & \multicolumn{2}{c}{LV $\cap$ RV} 
    & \multicolumn{2}{c}{RV $\cap$ RA}
    & \multicolumn{2}{c}{Average} 
    & \multicolumn{2}{c}{Average} 
    \\
\cmidrule(lr){2-3} \cmidrule(lr){4-5} \cmidrule(lr){6-7} \cmidrule(lr){8-9} \cmidrule(l){10-11}

\textbf{Method}
    & $\text{CD}\downarrow$ & $\text{NC}\uparrow$ 
    & $\text{CD}\downarrow$ & $\text{NC}\uparrow$ 
    & $\text{CD}\downarrow$ & $\text{NC}\uparrow$ 
    & $\text{CD}\downarrow$ & $\text{NC}\uparrow$ 
    & $\text{CD}\downarrow$ & $\text{NC}\uparrow$ 
    \\ 
\cmidrule(r){1-1} \cmidrule(lr){2-3} \cmidrule(lr){4-5} \cmidrule(lr){6-7} \cmidrule(lr){8-9} \cmidrule(l){10-11}

Ours
    & \cellcolor{blue!25}$0.7 \pm 0.3$ & \cellcolor{red!10}$0.95 \pm 0.01$
    & \cellcolor{blue!25}$1.4 \pm 0.7$ & \cellcolor{blue!25}$0.95 \pm 0.01$
    & \cellcolor{blue!25}$1.9 \pm 1.2$ & \cellcolor{red!10}$0.94 \pm 0.02$
    & \cellcolor{blue!25}$1.3 \pm 0.6$ & \cellcolor{blue!25}$0.95 \pm 0.01$
    & \cellcolor{blue!25}$1.0 \pm 0.3$ & \cellcolor{red!10}$0.93 \pm 0.01$
    \\

MedTet
    & $3.4 \pm 2.3$ & \cellcolor{blue!25}$0.96 \pm 0.01$
    & $5.6 \pm 3.4$ & \cellcolor{blue!25}$0.95 \pm 0.01$
    & \cellcolor{red!10}$2.7 \pm 2.0$ & \cellcolor{blue!25}$0.95 \pm 0.01$
    & $3.9 \pm 2.4$ & \cellcolor{blue!25}$0.95 \pm 0.01$
    & \cellcolor{red!10}$1.1 \pm 0.3$ & \cellcolor{blue!25}$0.94 \pm 0.01$
    \\

MeshDeformNet    
    & $2.4 \pm 1.5$ & $0.91 \pm 0.02$
    & $6.0 \pm 4.1$ & $0.90 \pm 0.05$
    & $7.5 \pm 9.5$ & $0.89 \pm 0.04$
    & $5.3 \pm 4.8$ & $0.90 \pm 0.03$
    & $2.5 \pm 1.4$ & $0.88 \pm 0.02$
    \\

nnU-Net 
    & \cellcolor{red!10}$0.9 \pm 0.4$ & $0.90 \pm 0.00$
    & \cellcolor{red!10}$1.6 \pm 0.6$ & $0.88 \pm 0.01$
    & $2.4 \pm 1.4$ & $0.89 \pm 0.01$
    & \cellcolor{red!10}$1.6 \pm 0.7$ & $0.89 \pm 0.01$
    & \cellcolor{red!10}$1.1 \pm 0.3$ & $0.87 \pm 0.01$
    \\

\hline
\end{tabular}
}
\end{table*}
\endgroup

\begingroup
\setlength{\tabcolsep}{4pt}
\begin{table}
\caption{Comparing Dice coefficients. \textit{Our meshes are watertight. After voxelization, \pf{which effectively reduces precision},  their volumetric accuracy is similar to that of nnU-Net.}}
\vspace{-3mm}
\label{tab:volume}
\centering
\renewcommand{\arraystretch}{1.2}
\resizebox{0.5\textwidth}{!}{
\begin{tabular}{@{}lccccccccc@{}}
\hline
& \multicolumn{5}{c}{\textbf{MM-WHS}} & \multicolumn{3}{c}{\textbf{MSD-Hippocampus}} \\
\cmidrule(lr){2-6} \cmidrule(l){7-9}
  
\textbf{Method} & \textbf{LV} & \textbf{RV} & \textbf{LA} & \textbf{RA} & \textbf{Mean} 
    & \textbf{Anterior} & \textbf{Posterior} & \textbf{Mean} \\

\cmidrule(r){1-1} \cmidrule(lr){2-6} \cmidrule(l){7-9}

Ours (Mesh) 
    & $0.95$ & $0.90$ & $0.92$ & $0.88$ & $0.91$ 
    & 0.87 & 0.85 & 0.86 \\

nnU-Net 
    & $0.95$ & $0.90$ & $0.92$ & $0.89$ & $0.92$
    & 0.88 & 0.86 & 0.87 \\

\hline
\end{tabular}
}
\end{table}
\endgroup
 
\parag{Heart Reconstruction.} 
In Tab.~\ref{tab:surface-heart}, we report per-class Chamfer Distance (CD) and Normal Consistency (NC) values for our method and the baselines on the MM-WHS-4 dataset both intersection classes---regions where anatomical structures meet---and other classes---the actual heart chambers. Lower CD and higher NC indicate better surface alignment and smoother geometry.

Our method achieves the best overall performance across the majority of intersection classes, yielding the lowest average Chamfer Distance and highest or near-highest Normal Consistency. Notably, it ranks first in 5 out of 6 CD columns and achieves excellent NC performance. This indicates that it not only captures accurate surfaces but also ensures smooth transitions in geometrically complex areas, which is critical when modeling realistic cardiac interfaces. In terms of volume reconstruction accuracy as reported in Tab.~\ref{tab:volume}, our method performs comparably to nnU-Net, which is unsurprising because the differences  between reconstructions occur mostly at the boundaries between substructures, which only represent a small fraction of the total number of voxels.  Furthermore,  its voxel-based nature results in visible artifacts  when turning the reconstructed volume into 3D surfaces. As can be seen in the third column of Fig.~\ref{fig:qualitative-heart}, nnU-Net outputs exhibit rough and uneven walls, especially in the "LV + Walls" and "LV (No Walls)" views. This translates into the comparatively lower NC score of Tab.~\ref{tab:volume}. Such artifacts do not exist in real human anatomy. In reality, organs like the heart have naturally smooth and continuous surfaces, shaped by biological processes and soft tissue mechanics. This problem is not specific to nnU-Net; it is inherent to fixed-resolution voxel-based representations.

{

\begin{figure}[t]
    \centering
    \newcommand{\imgheight}{1.95cm}
    \newcommand{\colw}{1.95cm}
    \newcommand{\imgpad}{\hspace{0pt}}
    \newcommand{\headerpad}{\vspace{2pt}}
    \newcommand{\rowpad}{\vspace{5pt}}
    \newcommand{\labelbox}[1]{%
        \raisebox{0.8cm}[0cm][-2cm]{%
            \makebox[0pt][r]{%
                \rotatebox[origin=c]{90}{\centering #1}%
            }%
        }%
    }

    \makebox[\colw][c]{\textbf{Ground Truth}} \imgpad
    \makebox[\colw][c]{\textbf{Ours}} \imgpad
    \makebox[\colw][c]{\textbf{nnU-Net}} \imgpad
    \makebox[\colw][c]{\textbf{MedTet}}\\[5pt]

    \labelbox{Whole Heart}
    \includegraphics[height=\imgheight]{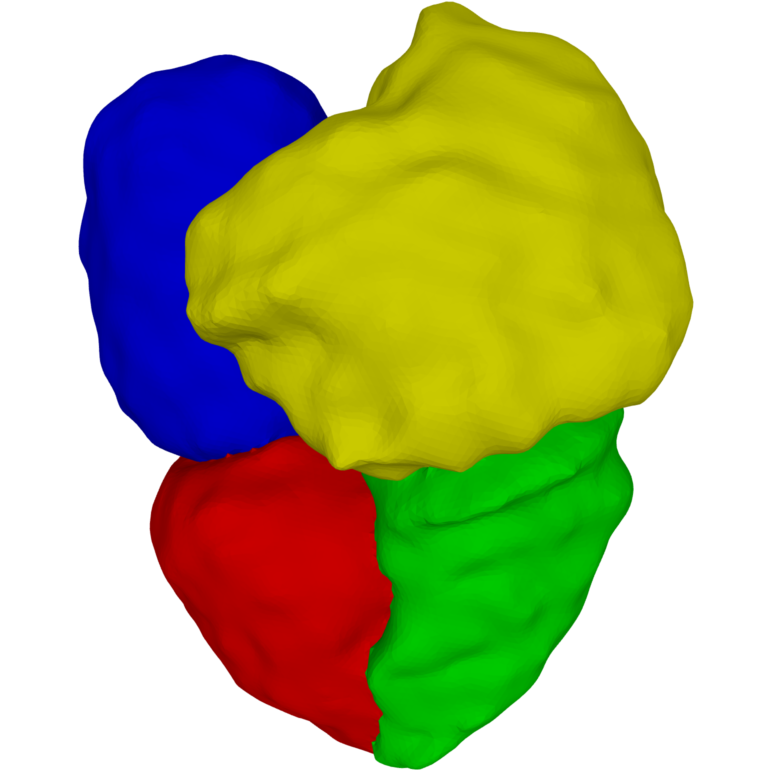}\imgpad
    \includegraphics[height=\imgheight]{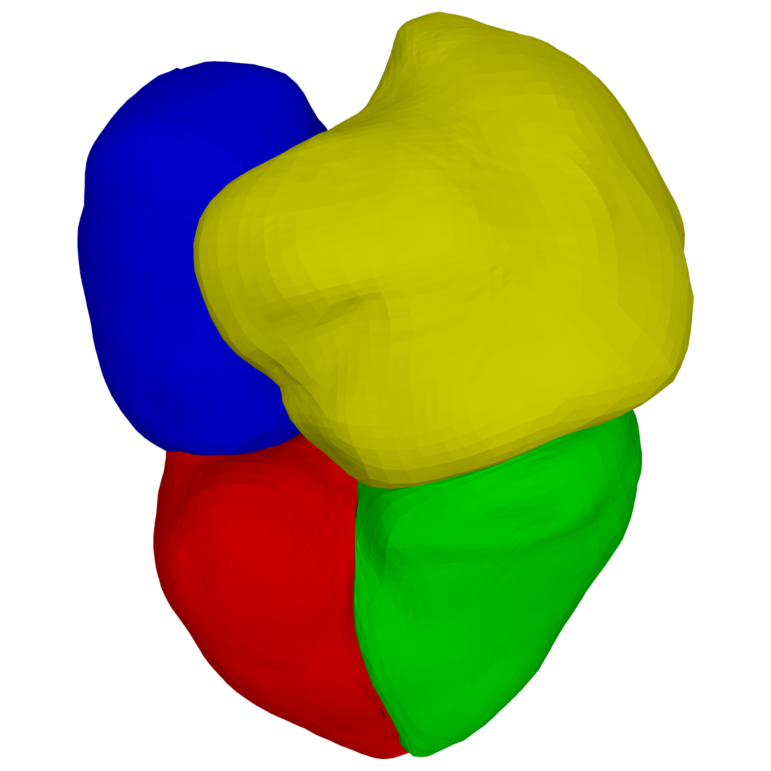}\imgpad
    \includegraphics[height=\imgheight]{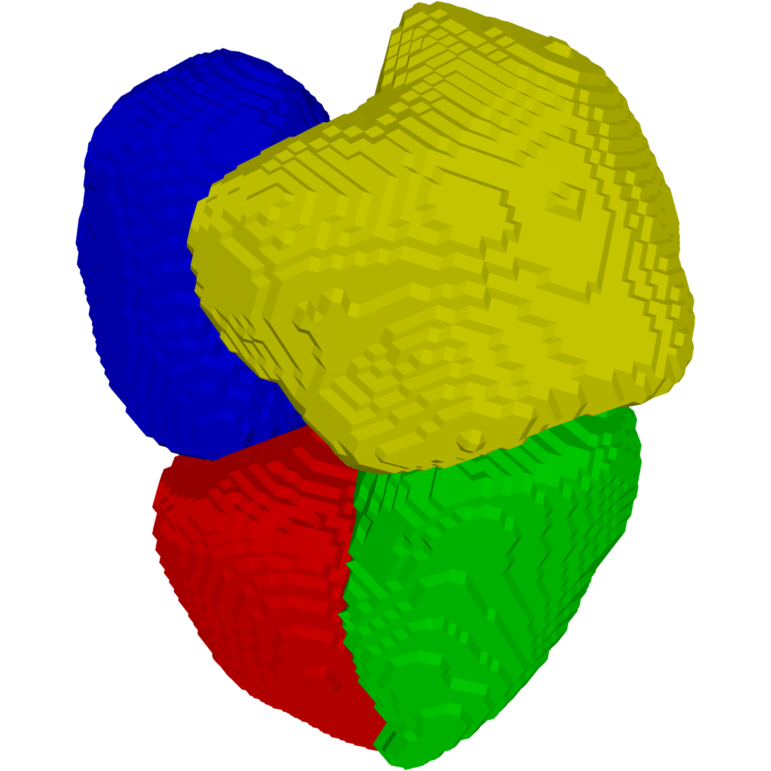}\imgpad
    \includegraphics[height=\imgheight]{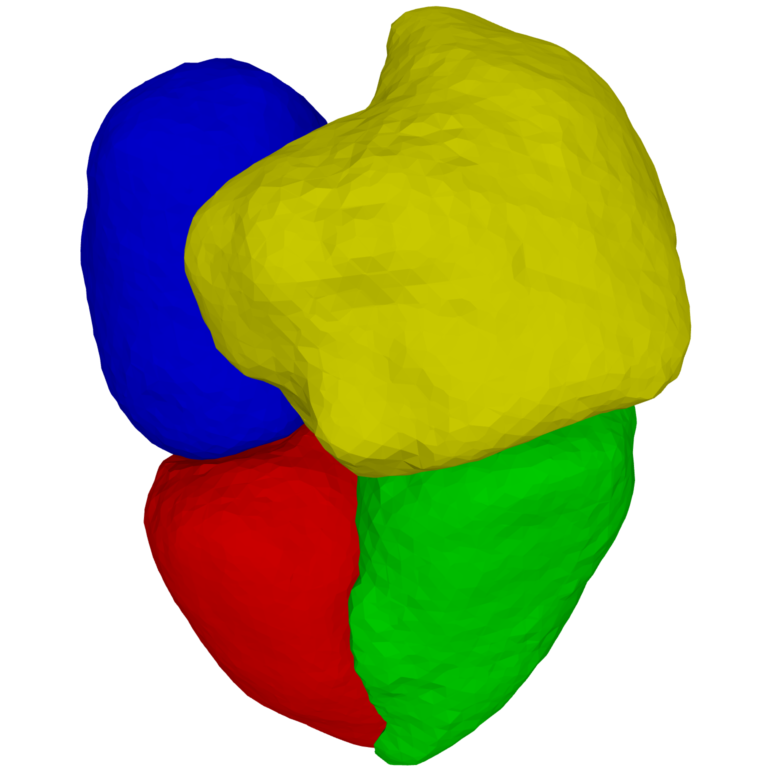}
    \rowpad

    \labelbox{LV + Walls}
    \includegraphics[height=\imgheight]{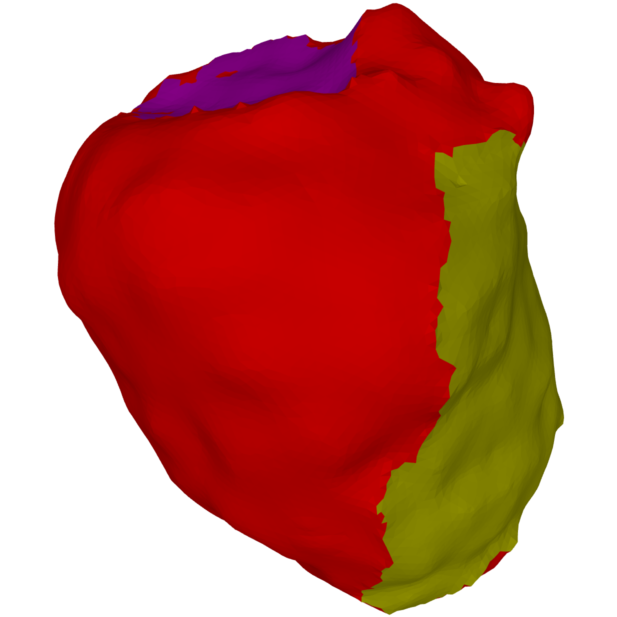}\imgpad
    \includegraphics[height=\imgheight]{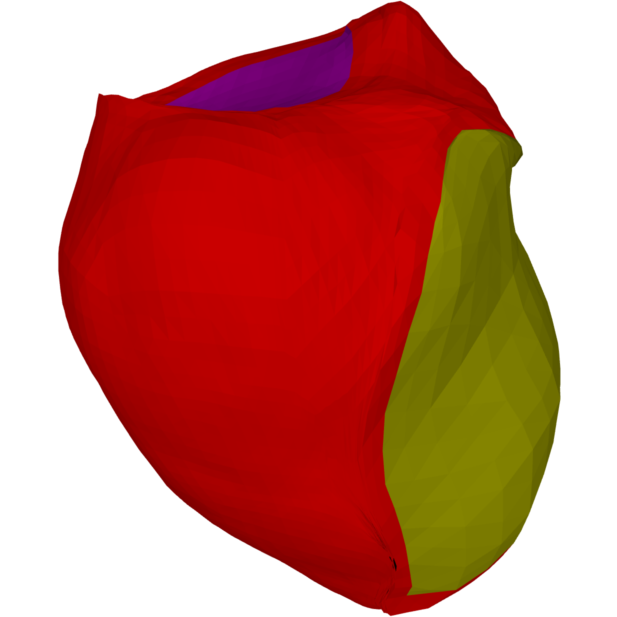}\imgpad
    \includegraphics[height=\imgheight]{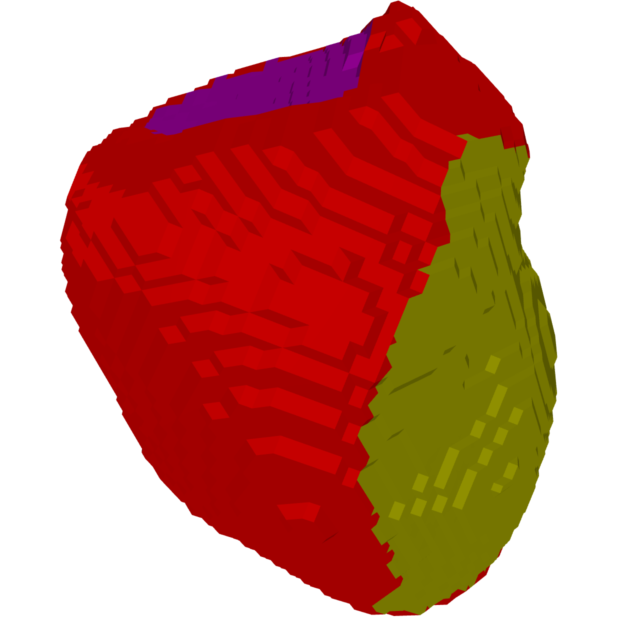}\imgpad
    \includegraphics[height=\imgheight]{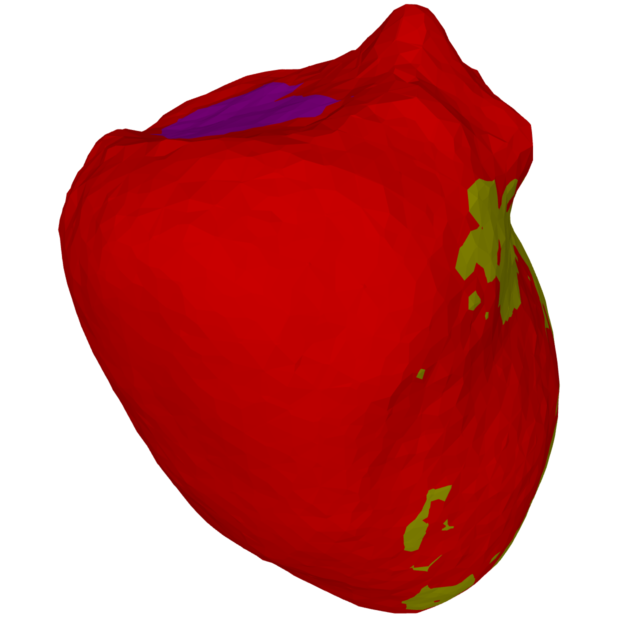}
    \rowpad

    \labelbox{LV w/o Walls}
    \includegraphics[height=\imgheight]{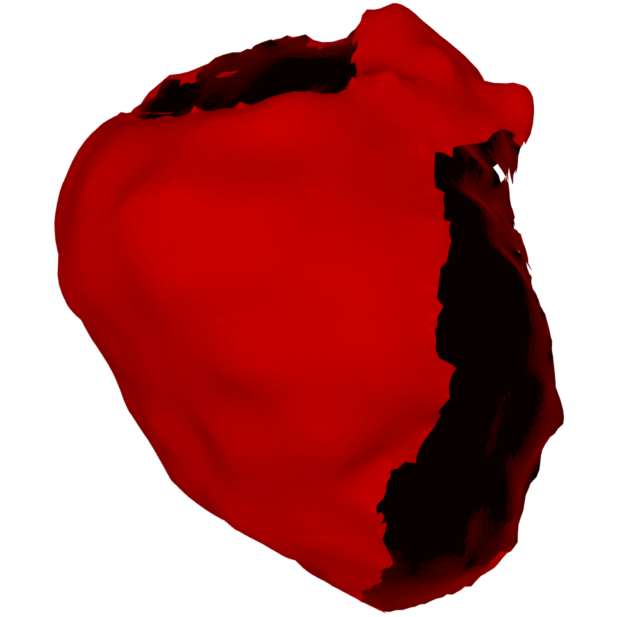}\imgpad
    \includegraphics[height=\imgheight]{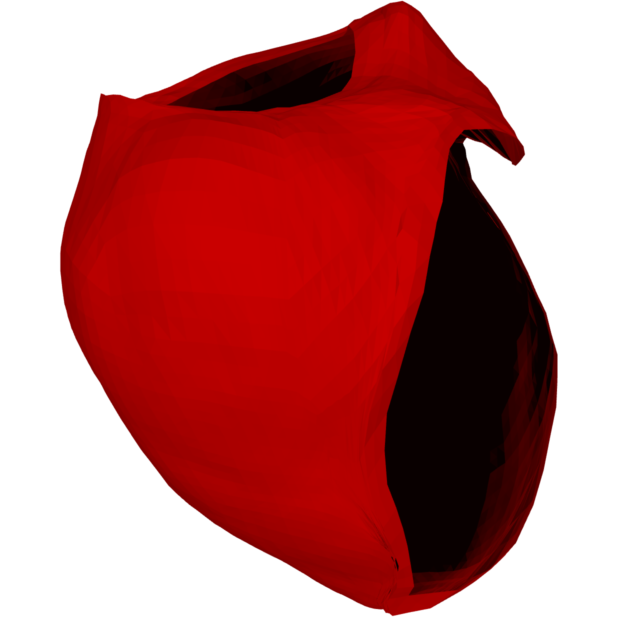}\imgpad
    \includegraphics[height=\imgheight]{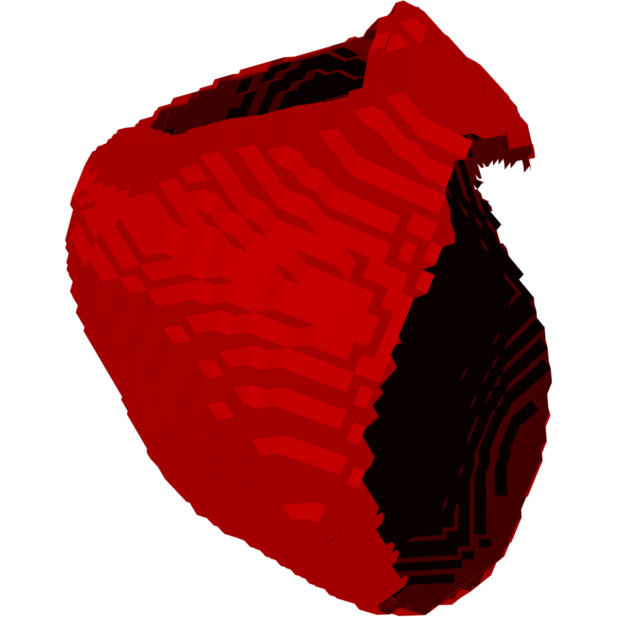}\imgpad
    \includegraphics[height=\imgheight]{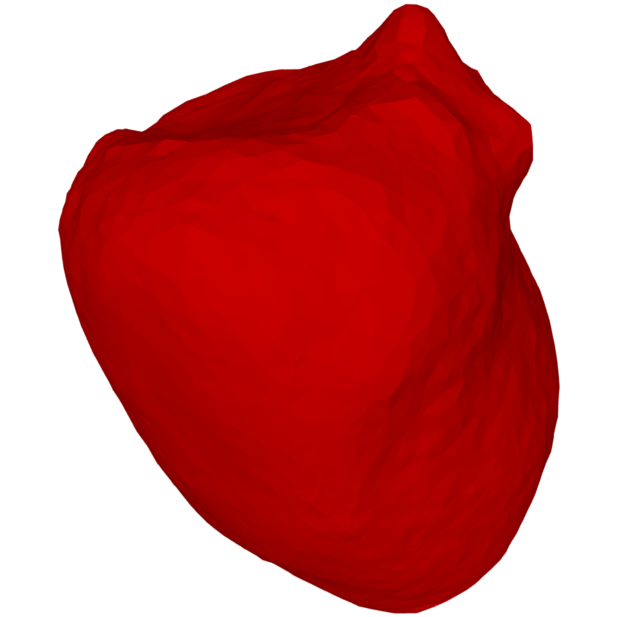}
    \rowpad

    \labelbox{LV-RV Wall}
    \includegraphics[height=\imgheight]{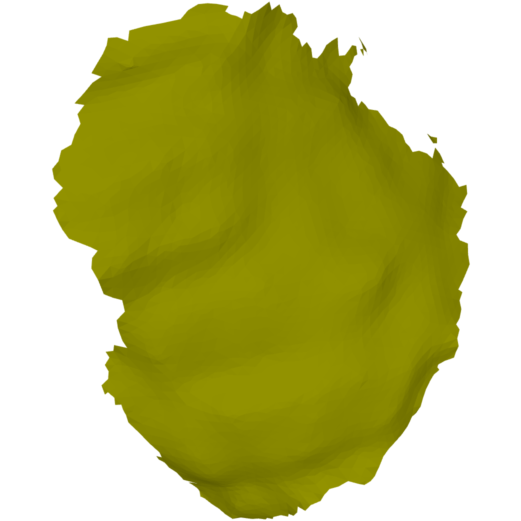}\imgpad
    \includegraphics[height=\imgheight]{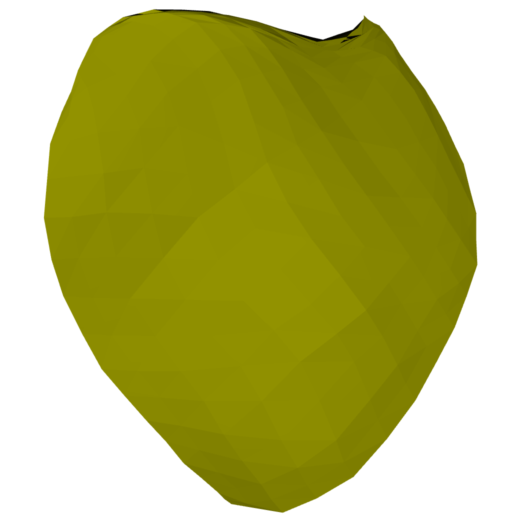}\imgpad
    \includegraphics[height=\imgheight]{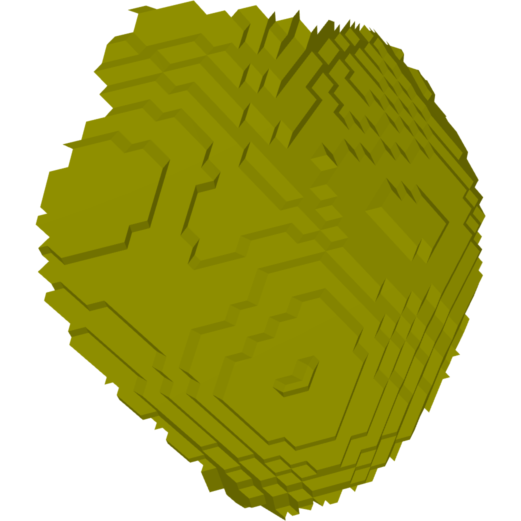}\imgpad
    \includegraphics[height=\imgheight]{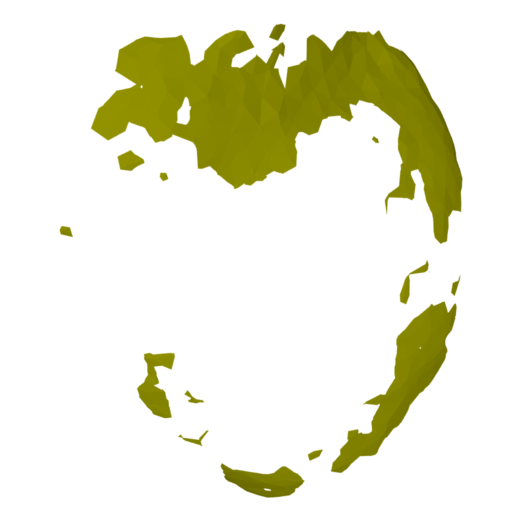}
    \rowpad

    \caption{\textbf{Qualitative comparison of 4-chamber heart outputs.}
    Left column describes the anatomical view. Each row compares methods visually. Our method produces smoother and more consistent geometry, especially in edge-sensitive regions.}
    \label{fig:qualitative-heart}
\end{figure}

}
 
Conversely, MedTet performs significantly worse than us in terms of CD score, especially in challenging regions such as LV-RV and RV-RA, but outperforms us slightly in terms of NC scores in the intersection classes, This suggests that while MedTet produces smooth normals, it is less good at precisely aligning surfaces.

These observations are further supported by the topological metrics presented in Tab.~\ref{tab:topo-heart}, which quantify structural correctness. Our method achieves zero intersection volume and no unwanted gaps, indicating clean, watertight surface reconstructions. In contrast, nnU-Net, while free from surface self-intersections, introduces an average of nearly one unwanted surface gap per case—further highlighting the challenges of voxel-based reconstruction in topologically complex regions. MedTet and MeshDeformNet introduce overlapping volumes between structures, and MeshDeformNet frequently outputs degenerate meshes, making a correct volume calculation difficult. The number of gaps is ill-defined for these because they are made of several independent connected components. 

In short, by jointly working with surface and volume representations, our method  produces smooth, coherent, and anatomically plausible geometry, while closely matching what we expect from actual biological structures. This is especially important in sensitive areas like the LV–RV wall, where precise surface detail matters both visually and functionally.

\begingroup
\begin{table}
\vspace{2mm}
\caption{Topology metrics on MM-WHS-4. }
\vspace{-3mm}
\label{tab:topo-heart}
\centering
\renewcommand{\arraystretch}{1.2}
\resizebox{0.5\textwidth}{!}{
\begin{tabular}{@{}lcc@{}}
\hline
\textbf{Method} & \textbf{Intersection Volume} & \textbf{\# of Unwanted Gaps} \\
\hline
Ours             & $\mathbf{0}$ & $\mathbf{0}$ \\
nnU-Net          & $\mathbf{0}$ & $0.85 \pm 0.02$ \\
MedTet           & $0.18\% \pm 0.02\%$ & undefined \\
MeshDeformNet    & degenerate meshes & undefined \\
\hline
\end{tabular}
}
\end{table}
\endgroup

\parag{Hippocampus and Lung Reconstruction.} 
We report our hippocampus reconstruction results in Tab.~\ref{tab:surface-hippo} with a focus on the junction between anterior and posterior segments, along with quantitative results in Fig.~\ref{fig:qualitative-hippo}. While all methods yield similar CD numbers, ours delivers significantly higher NC at this junction, indicating a smoother and more realistic transition. Our method's ability to model this boundary as a clean, topologically consistent surface—without jagged artifacts or mesh discontinuities—demonstrates its strength in capturing anatomically realistic intersections, even in small, structurally delicate regions. We observe similar behavior for lung reconstruction results, as can be seen in Tab.~\ref{tab:surface-lung}, with superior intersection reconstruction compared to mesh-based methods, and superior normal consistency compared to voxel segmentation.

\newcolumntype{?}{!{\vrule width 1pt}}

\begingroup
\setlength{\tabcolsep}{4pt}
\begin{table}
\caption{Per-class surface metrics on MSD-Hippocampus.}
\vspace{-3mm}
\label{tab:surface-hippo}
\centering
\renewcommand{\arraystretch}{1.2}
\resizebox{0.5\textwidth}{!}{
\begin{tabular}{@{}lcccccc}
\hline

& \multicolumn{2}{c}{Ant. $\cap$ Post.} 
& \multicolumn{2}{c}{Anterior}
& \multicolumn{2}{c}{Posterior}
    \\

\cmidrule(lr){2-3} \cmidrule(lr){4-5} \cmidrule(r){6-7}

& $\text{CD}\downarrow$ & $\text{NC}\uparrow$ 
    & $\text{CD}\downarrow$ & $\text{NC}\uparrow$ 
    & $\text{CD}\downarrow$ & $\text{NC}\uparrow$ 
    \\ 

\cmidrule(r){1-1} \cmidrule(lr){2-3} \cmidrule(lr){4-5} \cmidrule(r){6-7}

Ours
    & $\mathbf{3.3}$ & $\mathbf{0.98}$
    & $\mathbf{1.3}$ &$\mathbf{0.90}$ 
    & 1.3 & $\mathbf{0.90}$ 
    \\

nnU-Net 
    & 3.5 & $\mathbf{0.98}$
    & $\mathbf{1.3}$ & 0.80
    & $\mathbf{1.2}$ & 0.80
    \\

\hline
\end{tabular}
}
\end{table}
\endgroup

{
\newcommand{\imgwidth}{0.16\textwidth}
\newcommand{\imH}{3cm}
\newcommand{\spcc}{\hspace{0.025\textwidth}}
\newcommand{\rowheaderwidth}{3cm}

\newcommand{\wid}{2}
\newcommand{\wwid}{49}

\begin{figure}[th]
    \centering
    \renewcommand{\arraystretch}{0.1} %
    \setlength{\tabcolsep}{0pt}       %

    \begin{tabular}{
        >{\centering\arraybackslash}m{\imgwidth}%
        >{\centering\arraybackslash}m{\imgwidth}%
        >{\centering\arraybackslash}m{\imgwidth}%
    }
        \textbf{Ground Truth} & \textbf{Ours} & \textbf{nnU-Net} \\ 

        \includegraphics[width=\imgwidth,height=\imH]{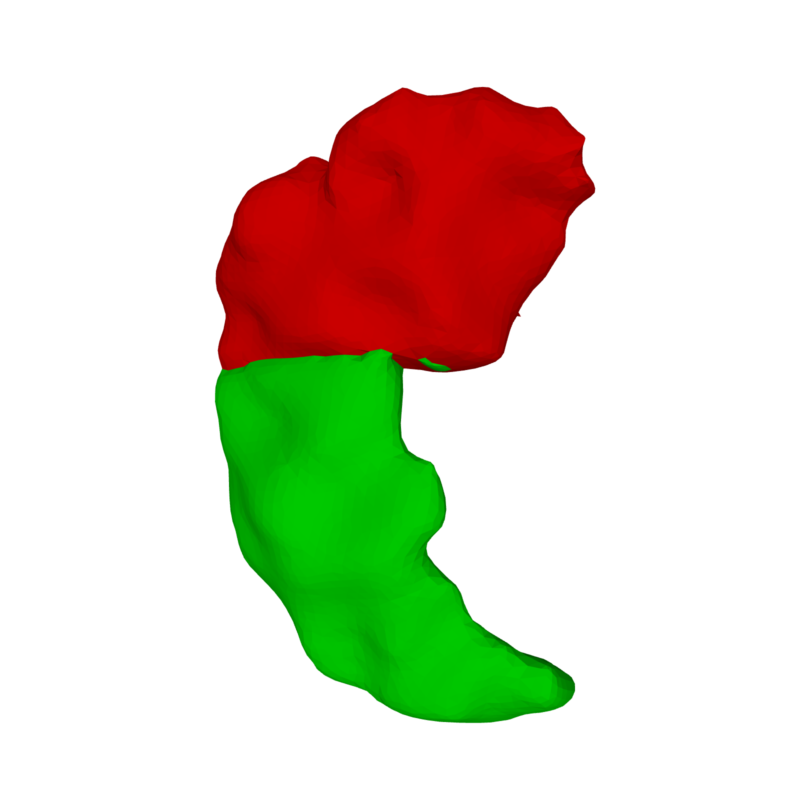} &
        \includegraphics[width=\imgwidth,height=\imH]{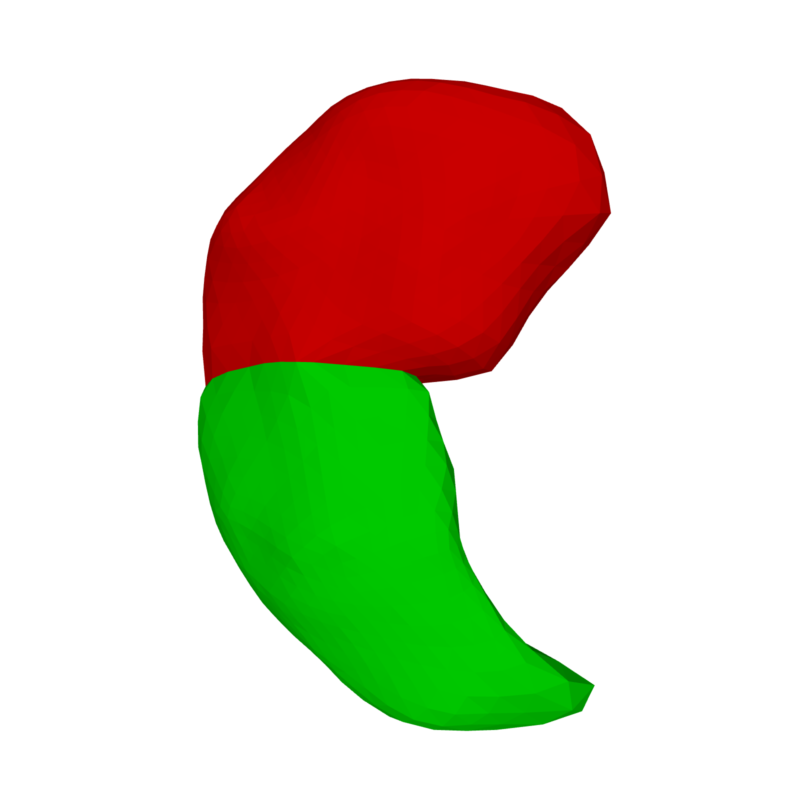} &
        \includegraphics[height=2.4cm]{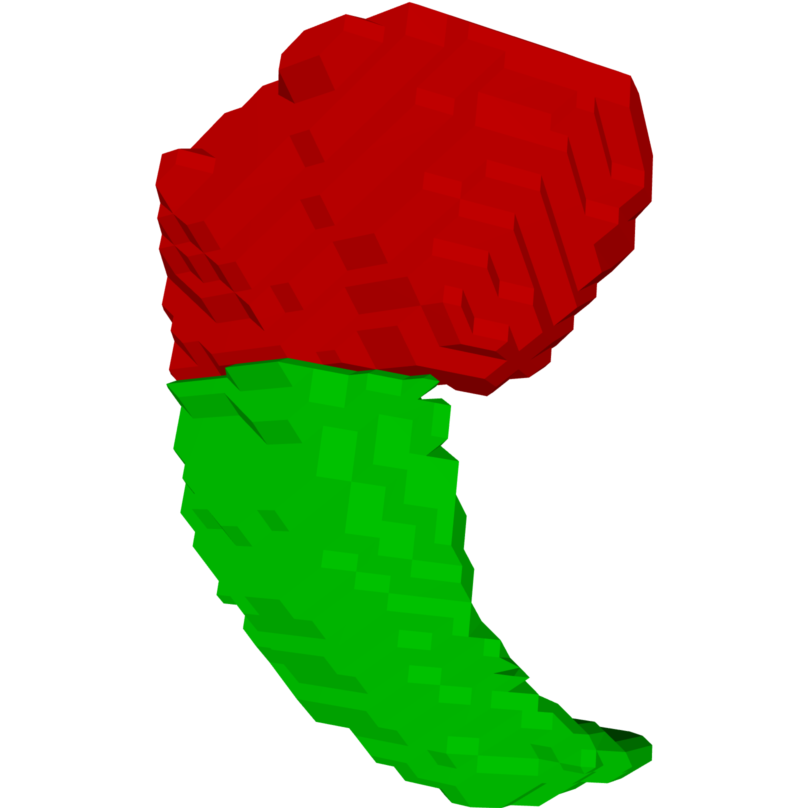}
        \\

        \includegraphics[width=\imgwidth]{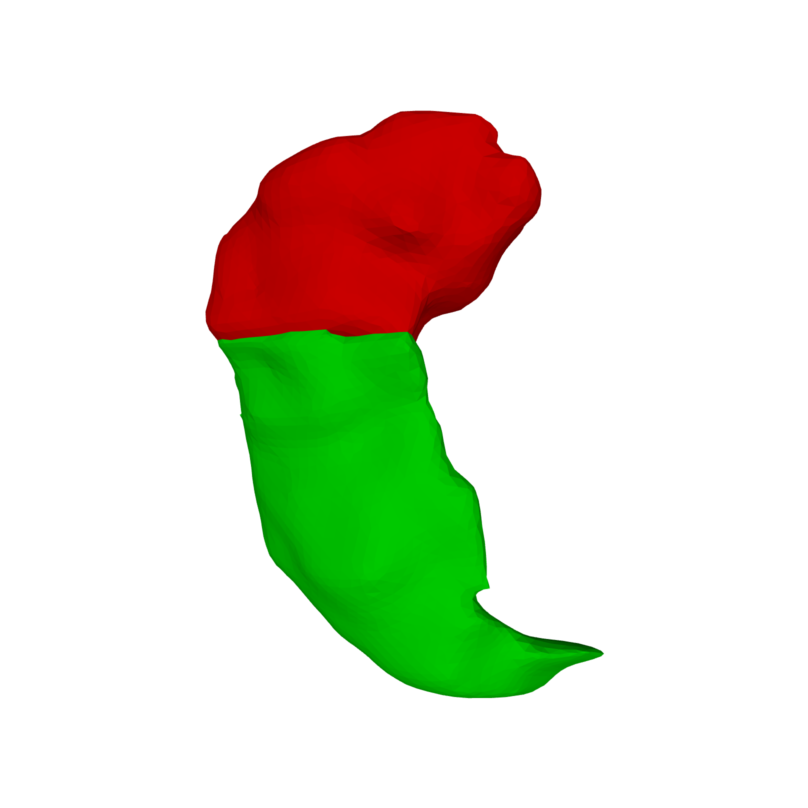} &
        \includegraphics[width=\imgwidth]{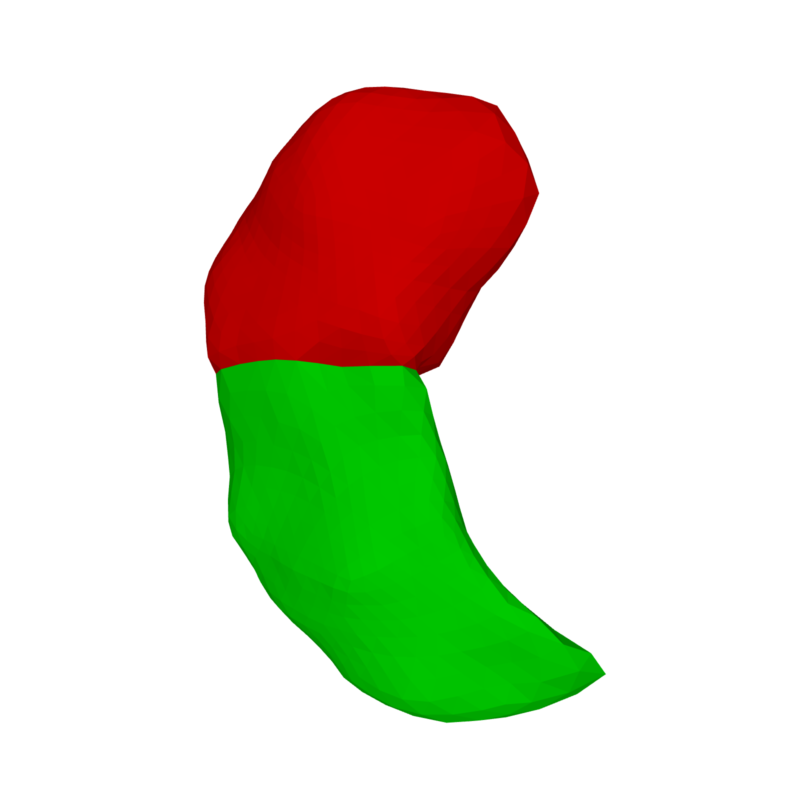} &
        \includegraphics[height=2.4cm]{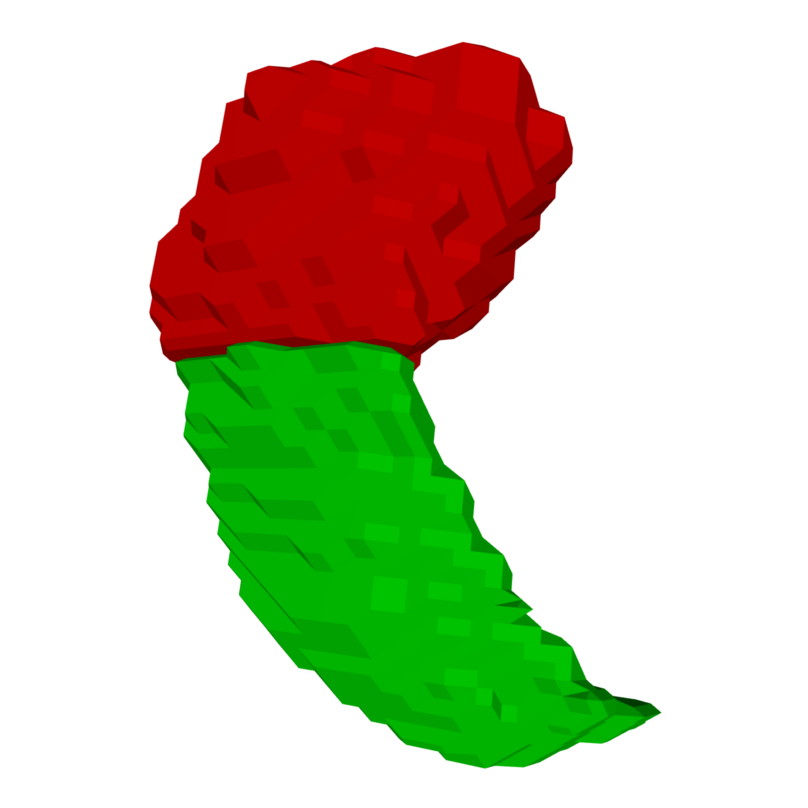}
        \\
    \end{tabular}
    \vspace{-4mm}
    \caption{\protect\textbf{Qualitative comparison of hippocampus outputs.} The ground truth remains irregular even after smoothing due to the low data resolution.}
    \label{fig:qualitative-hippo}
    \end{figure}
}

\vspace{3mm}
\begingroup
\begin{table}
\caption{Per-class surface metrics on TS-Lung.}
\vspace{-3mm}
\label{tab:surface-lung}
\centering
\setlength{\tabcolsep}{1pt}
\renewcommand{\arraystretch}{1.2}
\resizebox{0.5\textwidth}{!}{
\begin{tabular}{@{}lcccccccccccc}
\hline

& \multicolumn{10}{c}{\textbf{Intersection Classes}} & \multicolumn{2}{c}{\textbf{Other Classes}} \\

\cmidrule(lr){2-11} \cmidrule(l){12-13}

& \multicolumn{2}{c}{LR $\cap$ MR} 
    & \multicolumn{2}{c}{LR $\cap$ UR} 
    & \multicolumn{2}{c}{MR $\cap$ UR}
    & \multicolumn{2}{c}{LL $\cap$ UL}
    & \multicolumn{2}{c}{Average} 
    & \multicolumn{2}{c}{Average} 
    \\

\cmidrule(lr){2-3} \cmidrule(lr){4-5} \cmidrule(lr){6-7} \cmidrule(lr){8-9} \cmidrule(lr){10-11} \cmidrule(l){12-13}

\textbf{Method}
    & $\text{CD}\downarrow$ & $\text{NC}\uparrow$ 
    & $\text{CD}\downarrow$ & $\text{NC}\uparrow$ 
    & $\text{CD}\downarrow$ & $\text{NC}\uparrow$ 
    & $\text{CD}\downarrow$ & $\text{NC}\uparrow$ 
    & $\text{CD}\downarrow$ & $\text{NC}\uparrow$ 
    & $\text{CD}\downarrow$ & $\text{NC}\uparrow$ 
    \\ 

\cmidrule(r){1-1} \cmidrule(lr){2-3} \cmidrule(lr){4-5} \cmidrule(lr){6-7} \cmidrule(lr){8-9} \cmidrule(l){10-11}

Ours
    & $3.79$ & $\mathbf{0.96}$
    & $1.03$ & $\mathbf{0.98}$
    & $1.04$ & $\mathbf{0.96}$
    & $2.88$ & $\mathbf{0.94}$
    & $2.18$ & $\mathbf{0.96}$
    & $0.98$ & $\mathbf{0.95}$
    \\

nnU-Net 
    & $\mathbf{3.57}$ & $0.90$
    & $\mathbf{0.96}$ & $0.89$
    & $\mathbf{0.93}$ & $0.90$
    & $\mathbf{2.26}$ & $0.90$
    & $\mathbf{1.93}$ & $0.90$
    & $\mathbf{0.64}$ & $0.89$
    \\

\hline
\end{tabular}
}
\end{table}
\endgroup

\newcommand{\imgwidth}{0.16\textwidth}
\newcommand{\imH}{3cm}
\newcommand{\spcc}{\hspace{0.025\textwidth}}
\newcommand{\rowheaderwidth}{3cm}

\newcommand{\wid}{28}

\begin{figure}[th]
    \centering
    \renewcommand{\arraystretch}{1.5} %
    \setlength{\tabcolsep}{0pt}       %

    \begin{tabular}{
        >{\centering\arraybackslash}m{\imgwidth}%
        >{\centering\arraybackslash}m{\imgwidth}%
        >{\centering\arraybackslash}m{\imgwidth}%
    }
        \textbf{Ground Truth} & \textbf{Ours} & \textbf{nnU-Net} \\ 
        \vspace{1mm}

        \includegraphics[width=\imgwidth]{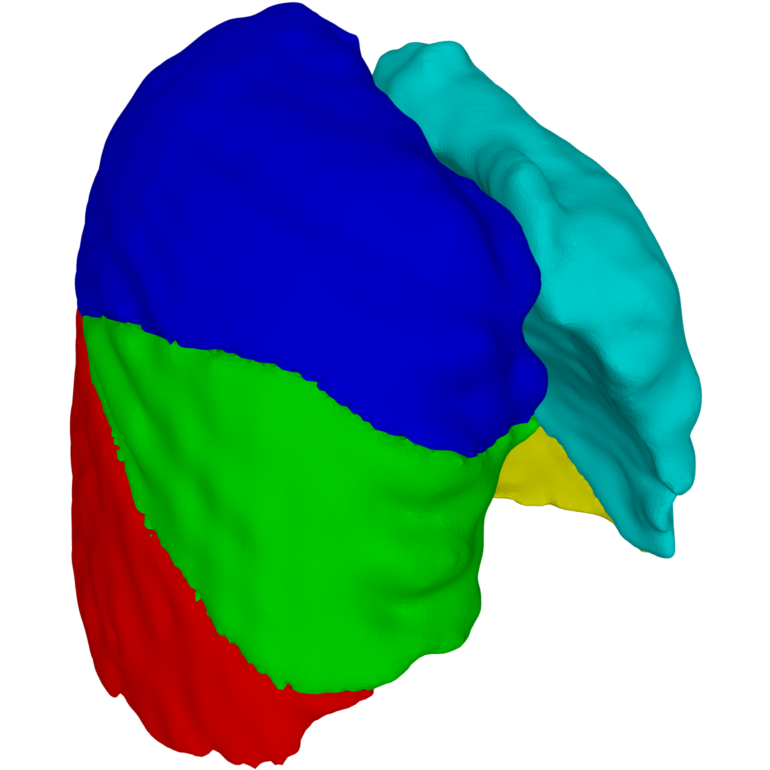} &
        \includegraphics[width=\imgwidth]{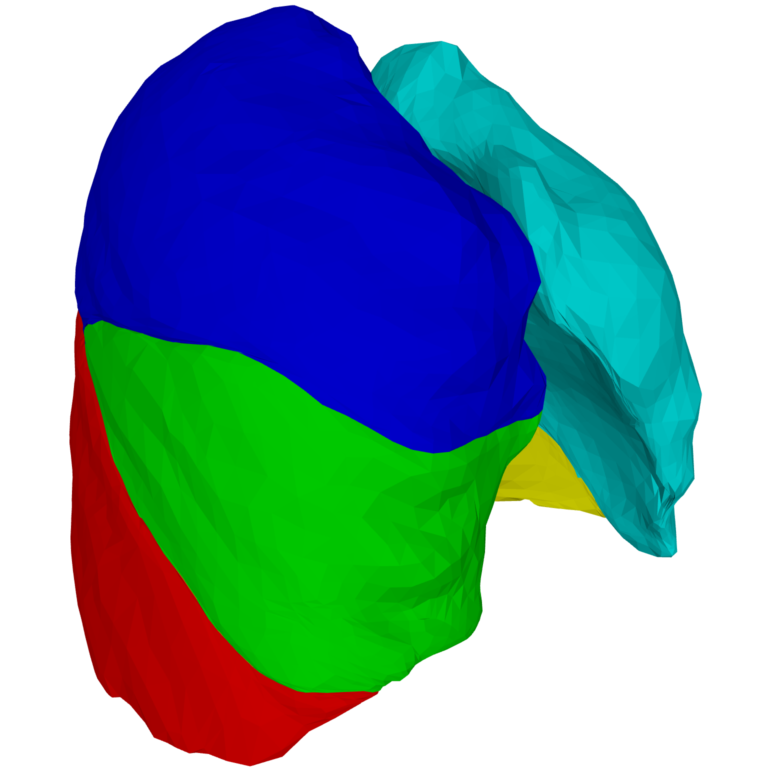} &
        \includegraphics[width=\imgwidth]{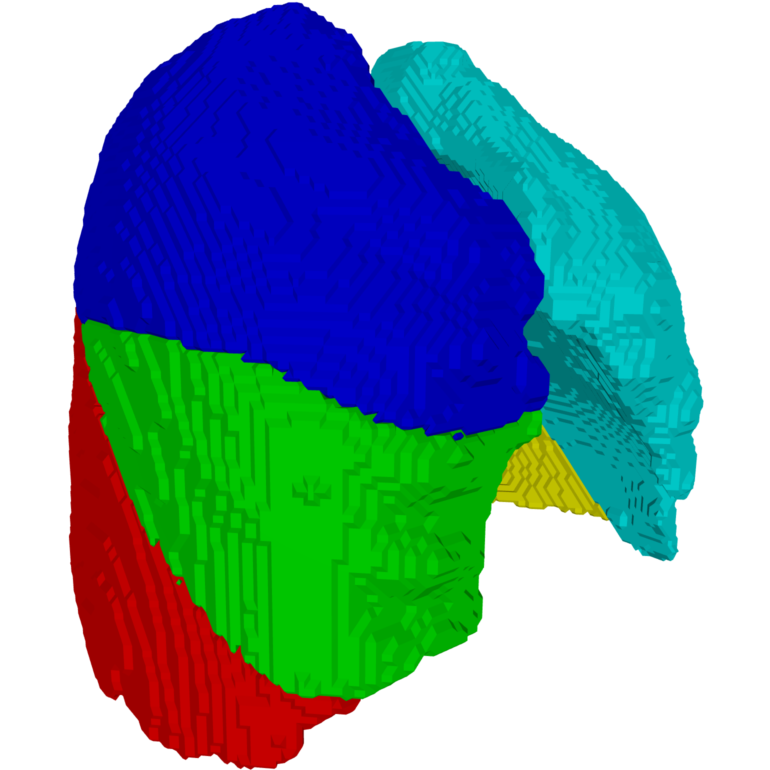}
        
    \end{tabular}
    \vspace{-3mm}
    \caption{\protect\textbf{Lung reconstructions.} \acro{} yields a smooth reconstruction free of residual and topological irregularities and with well-defined intersections. In contrast, the nnU-Net output is much rougher.}
    \label{fig:qualitative-lung}
\end{figure}

\vspace{-5mm}
\subsection{Effect of Training Set Size}

In medical applications, large training sets are not always available and being able to train a network using a relatively small dataset is desirable. With this in mind, we re-ran the hippocampus and lung reconstruction experiments, using only a fraction of the training database to train our approach and nnU-Net. We report the results in Fig.~\ref{fig:progressive}. In the low-data regime, our approach clearly outperforms nnU-net not only in NC terms but also in CD terms. This is due to the strong prior our approach imposes on the organ shape with the built-in template. While nnU-Net produces artifacts and disconnected regions in the low-data regime and greatly benefits from more data, our approach remains consistent even with few training samples.

\begin{figure}[th]
    \centering
    \subfloat[Lung reconstruction.\label{fig:lung-prog}]{
        \includegraphics[width=0.241\textwidth]{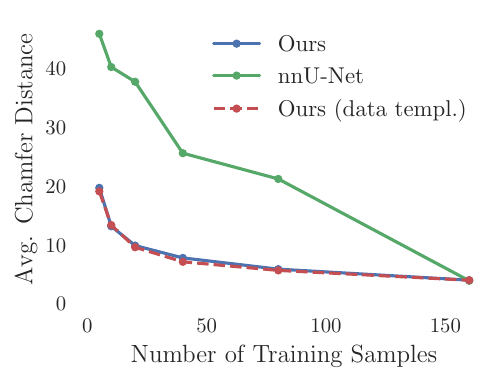}
    }
    \subfloat[Hippocampus reconstruction.\label{fig:hippo-prog}]{
        \includegraphics[width=0.241\textwidth]{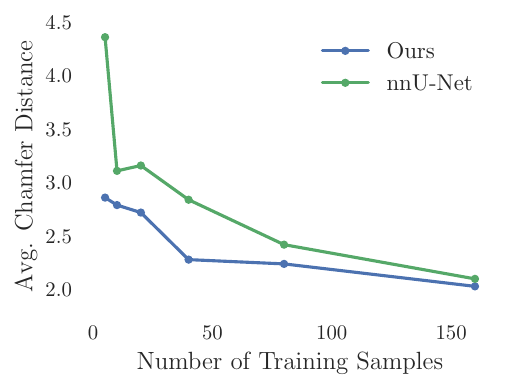}
    }
    \vspace{-3mm}
    \caption{\protect\textbf{Performance in the low-data regime.} With few training samples available, our approach yields more accurate surface reconstructions than nnU-Net. We also show the reconstruction performance of using a data template for the lung, which is very similar to the performance of using a hand-made template.}
    \label{fig:progressive}
\end{figure}

\vspace{-5mm}
\section{Conclusion}

We introduced \acro{}, a unified framework for jointly reconstructing multiple organ components. Unlike existing surface-based methods that model each component using independent, potentially overlapping meshes, \acro{} deforms a single multi-part template that preserves topology while accurately modeling shared interfaces. This enables state-of-the-art reconstruction accuracy with guaranteed topological consistency.

Our approach moves toward fully automatic generation of anatomically correct models suitable for downstream simulations without manual post-processing. Future work will focus on automating template creation to handle more complex anatomies, such as incorporating the aorta and pulmonary artery and capturing their temporal deformations.

 {
    \small
    \bibliographystyle{ieeenat_fullname}

\begin{thebibliography}{40}
\providecommand{\natexlab}[1]{#1}
\providecommand{\url}[1]{\texttt{#1}}
\expandafter\ifx\csname urlstyle\endcsname\relax
  \providecommand{\doi}[1]{doi: #1}\else
  \providecommand{\doi}{doi: \begingroup \urlstyle{rm}\Url}\fi

\bibitem[Antonelli et~al.(2022)Antonelli, Reinke, Bakas, Farahani,
  Kopp-Schneider, Landman, Litjens, Menze, Ronneberger, Summers,
  et~al.]{Antonelli2022}
Michela Antonelli, Annika Reinke, Spyridon Bakas, Keyvan Farahani, Annette
  Kopp-Schneider, Bennett~A Landman, Geert Litjens, Bjoern Menze, Olaf
  Ronneberger, Ronald~M Summers, et~al.
\newblock {The Medical Segmentation Decathlon}.
\newblock \emph{Nature communications}, 13\penalty0 (1):\penalty0 4128, 2022.

\bibitem[Bai et~al.(2013)Bai, Shi, O'regan, Tong, Wang, Jamil-Copley, Peters,
  and Rueckert]{Bai13b}
Wenjia Bai, Wenzhe Shi, Declan~P O'regan, Tong Tong, Haiyan Wang, Shahnaz
  Jamil-Copley, Nicholas~S Peters, and Daniel Rueckert.
\newblock {A Probabilistic Patch-Based Label Fusion Model for Multi-Atlas
  Segmentation with Registration Refinement: Application to Cardiac MRI
  Images}.
\newblock \emph{IEEE Transactions on Medical Imaging}, 32\penalty0
  (7):\penalty0 1302--1315, 2013.

\bibitem[Bertels et~al.(2019)Bertels, Eelbode, Berman, Vandermeulen, Maes,
  Bisschops, and Blaschko]{Bertels19a}
J. Bertels, T. Eelbode, M. Berman, D. Vandermeulen, F. Maes, R. Bisschops, and
  M. Blaschko.
\newblock {Optimizing the Dice Score and Jaccard Index for Medical Image
  Segmentation: Theory and Practice}.
\newblock In \emph{Conference on Medical Image Computing and Computer Assisted
  Intervention}, 2019.

\bibitem[Bongratz et~al.(2022)Bongratz, Rickmann, P{\"o}lsterl, and
  Wachinger]{Bongratz22}
F. Bongratz, A.-M. Rickmann, S. P{\"o}lsterl, and C. Wachinger.
\newblock {Vox2cortex: Fast Explicit Reconstruction of Cortical Surfaces from
  3D Mri Scans with Geometric Deep Neural Networks}.
\newblock In \emph{Conference on Computer Vision and Pattern Recognition},
  2022.

\bibitem[Charton et~al.(2021)Charton, Baek, and Kim]{Charton21}
Jerome Charton, Stephen Baek, and Youngjun Kim.
\newblock {Mesh Repairing using Topology Graphs}.
\newblock \emph{Journal of Computational Design and Engineering}, 2021.

\bibitem[Chen et~al.(2024)Chen, Yang, Mercadier, Le, and Fua]{Chen24f}
Y. Chen, J. Yang, D.~S. Mercadier, H. Le, and P. Fua.
\newblock {Medtet: An Online Motion Model for 4D Heart Reconstruction}.
\newblock In \emph{arXiv Preprint}, 2024.

\bibitem[Desbrun et~al.(1999)Desbrun, Meyer, Schr\"{o}der, and Barr]{Desbrun99}
M. Desbrun, M. Meyer, P. Schr\"{o}der, and A.~H. Barr.
\newblock {Implicit Fairing of Irregular Meshes Using Diffusion and Curvature
  Flow}.
\newblock In \emph{ACM SIGGRAPH}, 1999.

\bibitem[Dice(1945)]{Dice45}
R.L. Dice.
\newblock {Measures of the Amount of Ecologic Association Between Species}.
\newblock \emph{{Ecology}}, 1945.

\bibitem[Falk et~al.(2018)Falk, Mai, Bensch, Çiçek, Abdulkadir, Marrakchi,
  B{\"o}hm, Deubner, J{\"a}ckel, Seiwald, Dovzhenko, Tietz, Bosco, Walsh,
  Saltukoglu, Tay, Prinz, Palme, Simons, Diester, Brox, and
  Ronneberger]{Falk18}
T. Falk, D. Mai, R. Bensch, {\"O}zg{\"u}n Çiçek, A. Abdulkadir, Y. Marrakchi,
  A. B{\"o}hm, J. Deubner, Z. J{\"a}ckel, K. Seiwald, A. Dovzhenko, O. Tietz,
  C.~D. Bosco, S. Walsh, D. Saltukoglu, T.~L. Tay, M. Prinz, K. Palme, M.
  Simons, I. Diester, T. Brox, and O. Ronneberger.
\newblock {U-Net: Deep Learning for Cell Counting, Detection, and Morphometry}.
\newblock \emph{Nature Methods}, 16:\penalty0 67--70, 2018.

\bibitem[Fischl(2012)]{Fischl12}
B. Fischl.
\newblock {Freesurfer}.
\newblock \emph{Neuroimage}, 62\penalty0 (2):\penalty0 774--781, 2012.

\bibitem[Gkioxari et~al.(2019)Gkioxari, Malik, and Johnson]{Gkioxari19}
G. Gkioxari, J. Malik, and J. Johnson.
\newblock {Mesh R-CNN}.
\newblock In \emph{International Conference on Computer Vision}, 2019.

\bibitem[Iglesias and Sabuncu(2015)]{Iglesias15}
Juan~Eugenio Iglesias and Mert~R Sabuncu.
\newblock {Multi-Atlas Segmentation of Biomedical Images: a Survey}.
\newblock \emph{Medical Image Analysis}, 24\penalty0 (1):\penalty0 205--219,
  2015.

\bibitem[Isensee et~al.(2021)Isensee, Jaeger, Kohl, Petersen, and
  Maier-Hein]{Isensee21}
F. Isensee, P.F. Jaeger, S.A.A. Kohl, J. Petersen, and K.H. Maier-Hein.
\newblock {Nnu-Net: A Self-Configuring Method for Deep Learning-Based
  Biomedical Image Segmentation}.
\newblock \emph{Nature Methods}, 18\penalty0 (2):\penalty0 203--211, 2021.

\bibitem[Isensee et~al.(2024)Isensee, Wald, Ulrich, Baumgartner, Roy,
  Maier-Hein, and Jaeger]{Isensee24}
F. Isensee, T. Wald, C. Ulrich, M. Baumgartner, S. Roy, K. Maier-Hein, and P.
  Jaeger.
\newblock {Nnu-Net Revisited: A Call for Rigorous Validation in 3D Medical
  Image Segmentation}.
\newblock In \emph{Conference on Medical Image Computing and Computer Assisted
  Intervention}, 2024.

\bibitem[Kepler(1619)]{Kepler19}
J. Kepler.
\newblock \emph{{Harmonice Mundi}}.
\newblock Linz, 1619.

\bibitem[Kong and Shadden(2021)]{Kong21b}
F. Kong and S.~C. Shadden.
\newblock {Whole Heart Mesh Generation for Image-Based Computational
  Simulations by Learning Free-From Deformations}.
\newblock In \emph{Conference on Medical Image Computing and Computer Assisted
  Intervention}, pages 550--559, 2021.

\bibitem[Kong et~al.(2021)Kong, Wilson, and Shadden]{Kong21a}
F. Kong, N. Wilson, and S.~C. Shadden.
\newblock {A Deep-Learning Approach for Direct Whole-Heart Mesh
  Reconstruction}.
\newblock \emph{Medical Image Analysis}, 74, 2021.

\bibitem[Le et~al.(2023)Le, , Talabot, Yang, and Fua]{Le23a}
Hieu Le, , Nicolas Talabot, Jiancheng Yang, and Pascal Fua.
\newblock {Enforcing Topological Interaction Between Implicit Surfaces via
  Uniform Sampling}.
\newblock In \emph{arXiv Preprint}, 2023.

\bibitem[Li et~al.(2024)Li, Camps, Rodriguez, and Grau]{li2024solving}
Lei Li, Julia Camps, Blanca Rodriguez, and Vicente Grau.
\newblock Solving the inverse problem of electrocardiography for cardiac
  digital twins: A survey.
\newblock \emph{IEEE Reviews in Biomedical Engineering}, 2024.

\bibitem[Lorensen and Cline(1987)]{Lorensen87}
W.E. Lorensen and H.E. Cline.
\newblock {Marching Cubes: {A} High Resolution 3{D} Surface Construction
  Algorithm}.
\newblock In \emph{ACM SIGGRAPH}, pages 163--169, 1987.

\bibitem[Pak et~al.(2024)Pak, Liu, Kim, Ozturk, Mckay, Roche, Gleason, and
  Duncan]{pak2024robust}
Daniel~H Pak, Minliang Liu, Theodore Kim, Caglar Ozturk, Raymond Mckay, Ellen~T
  Roche, Rudolph Gleason, and James~S Duncan.
\newblock Robust automated calcification meshing for personalized
  cardiovascular biomechanics.
\newblock \emph{NPJ Digital Medicine}, 7\penalty0 (1):\penalty0 213, 2024.

\bibitem[Qian et~al.(2025)Qian, Ugurlu, Fairweather, Toso, Deng, Strocchi,
  Cicci, Jones, Zaidi, Prasad, et~al.]{qian2025developing}
Shuang Qian, Devran Ugurlu, Elliot Fairweather, Laura~Dal Toso, Yu Deng, Marina
  Strocchi, Ludovica Cicci, Richard~E Jones, Hassan Zaidi, Sanjay Prasad,
  et~al.
\newblock Developing cardiac digital twin populations powered by machine
  learning provides electrophysiological insights in conduction and
  repolarization.
\newblock \emph{Nature Cardiovascular Research}, 4\penalty0 (5):\penalty0
  624--636, 2025.

\bibitem[Qiao et~al.(2025)Qiao, McGurk, Wang, Matthews, O’Regan, and
  Bai]{qiao2025personalized}
Mengyun Qiao, Kathryn~A McGurk, Shuo Wang, Paul~M Matthews, Declan~P O’Regan,
  and Wenjia Bai.
\newblock A personalized time-resolved 3d mesh generative model for unveiling
  normal heart dynamics.
\newblock \emph{Nature Machine Intelligence}, pages 1--12, 2025.

\bibitem[Ravi et~al.(2020)Ravi, Reizenstein, Novotny, Gordon, Lo, Johnson, and
  Gkioxari]{Ravi20}
Nikhila Ravi, Jeremy Reizenstein, David Novotny, Taylor Gordon, Wan-Yen Lo,
  Justin Johnson, and Georgia Gkioxari.
\newblock Accelerating 3d deep learning with pytorch3d.
\newblock \emph{arXiv Preprint}, 2020.

\bibitem[Ronneberger et~al.(2015)Ronneberger, Fischer, and Brox]{Ronneberger15}
O. Ronneberger, P. Fischer, and T. Brox.
\newblock {{U-Net}: Convolutional Networks for Biomedical Image Segmentation}.
\newblock In \emph{Conference on Medical Image Computing and Computer Assisted
  Intervention}, pages 234--241, 2015.

\bibitem[Sakata et~al.(2024)Sakata, Bradley, Prakosa, Yamamoto, Ali, Loeffler,
  Tice, Boyle, Kholmovski, Yadav, et~al.]{sakata2024assessing}
Kensuke Sakata, Ryan~P Bradley, Adityo Prakosa, Carolyna~AP Yamamoto,
  Syed~Yusuf Ali, Shane Loeffler, Brock~M Tice, Patrick~M Boyle, Eugene~G
  Kholmovski, Ritu Yadav, et~al.
\newblock Assessing the arrhythmogenic propensity of fibrotic substrate using
  digital twins to inform a mechanisms-based atrial fibrillation ablation
  strategy.
\newblock \emph{Nature cardiovascular research}, 3\penalty0 (7):\penalty0
  857--868, 2024.

\bibitem[Sakata et~al.(2025)Sakata, Yamamoto, Prakosa, Tice, Ali, Loeffler,
  Kholmovski, Sinha, Marine, Calkins, et~al.]{sakata2025digital}
Kensuke Sakata, Carolyna~AP Yamamoto, Adityo Prakosa, Brock~M Tice, Syed~Yusuf
  Ali, Shane Loeffler, Eugene~G Kholmovski, Sunil~Kumar Sinha, Joseph~E Marine,
  Hugh Calkins, et~al.
\newblock Digital twins enable stratification of persistent atrial fibrillation
  patients for ablation diminishing unnecessary heart damage.
\newblock \emph{npj Digital Medicine}, 8\penalty0 (1):\penalty0 256, 2025.

\bibitem[Salvador et~al.(2024)Salvador, Strocchi, Regazzoni, Augustin, Dede’,
  Niederer, and Quarteroni]{salvador2024whole}
Matteo Salvador, Marina Strocchi, Francesco Regazzoni, Christoph~M Augustin,
  Luca Dede’, Steven~A Niederer, and Alfio Quarteroni.
\newblock Whole-heart electromechanical simulations using latent neural
  ordinary differential equations.
\newblock \emph{NPJ Digital Medicine}, 7\penalty0 (1):\penalty0 90, 2024.

\bibitem[She et~al.(2023)She, Zhang, Zhang, Li, Yan, and Sun]{She23}
D. She, Y. Zhang, Z. Zhang, H. Li, Z. Yan, and X. Sun.
\newblock {Eoformer: Edge-Oriented Transformer for Brain Tumor Segmentation}.
\newblock In \emph{Conference on Medical Image Computing and Computer Assisted
  Intervention}, 2023.

\bibitem[Shen et~al.(2021)Shen, Gao, Yin, Liu, and Fidler]{Shen21a}
T. Shen, J. Gao, K. Yin, M.-Y. Liu, and S. Fidler.
\newblock {Deep Marching Tetrahedra: A Hybrid Representation for
  High-Resolution 3D Shape Synthesis}.
\newblock In \emph{Advances in Neural Information Processing Systems}, 2021.

\bibitem[Updegrove et~al.(2017)Updegrove, Wilson, Merkow, Lan, Marsden, and
  Shadden]{updegrove2017simvascular}
Adam Updegrove, Nathan~M Wilson, Jameson Merkow, Hongzhi Lan, Alison~L Marsden,
  and Shawn~C Shadden.
\newblock Simvascular: an open source pipeline for cardiovascular simulation.
\newblock \emph{Annals of biomedical engineering}, 45\penalty0 (3):\penalty0
  525--541, 2017.

\bibitem[Verh{\"u}lsdonk et~al.(2024)Verh{\"u}lsdonk, Grandits, Costabal,
  Pinetz, Krause, Auricchio, Haase, Pezzuto, and Effland]{Verhulsdonk24}
J. Verh{\"u}lsdonk, T. Grandits, F.~S. Costabal, T. Pinetz, R. Krause, A.
  Auricchio, G. Haase, S. Pezzuto, and A. Effland.
\newblock {Shape of My Heart: Cardiac Models through Learned Signed Distance
  Functions}.
\newblock In \emph{Medical Imaging with Deep Learning}, 2024.

\bibitem[Wasserthal et~al.(2023)Wasserthal, Breit, Meyer, Pradella, Hinck,
  Sauter, Heye, Boll, Cyriac, Yang, et~al.]{Wasserthal23}
Jakob Wasserthal, Hanns-Christian Breit, Manfred~T Meyer, Maurice Pradella,
  Daniel Hinck, Alexander~W Sauter, Tobias Heye, Daniel~T Boll, Joshy Cyriac,
  Shan Yang, et~al.
\newblock {Totalsegmentator: Robust Segmentation of 104 Anatomic Structures in
  CT Images}.
\newblock \emph{Radiology: Artificial Intelligence}, 2023.

\bibitem[Wickramasinghe et~al.(2020)Wickramasinghe, Remelli, Knott, and
  Fua]{Wickramasinghe20}
U. Wickramasinghe, E. Remelli, G. Knott, and P. Fua.
\newblock {Voxel2mesh: 3D Mesh Model Generation from Volumetric Data}.
\newblock In \emph{Conference on Medical Image Computing and Computer Assisted
  Intervention}, 2020.

\bibitem[Yang et~al.(2023)Yang, Tam, and Tang]{Yang23b}
H. Yang, R. Tam, and X. Tang.
\newblock {Whole-Heart Reconstruction with Explicit Topology Integrated
  Learning}.
\newblock In \emph{Conference on Medical Image Computing and Computer Assisted
  Intervention}, pages 106--115, 2023.

\bibitem[Yang et~al.(2024)Yang, Sedykh, Adhinarta, Le, and Fua]{Yang24a}
J. Yang, E. Sedykh, J. Adhinarta, H. Le, and P. Fua.
\newblock {Generating Anatomically Accurate Heart Structures via Neural
  Implicit Fields}.
\newblock In \emph{Conference on Medical Image Computing and Computer Assisted
  Intervention}, 2024.

\bibitem[Yang et~al.(2018)Yang, Pan, Amert, Wang, Yu, Berg, and Lin]{Yang18f}
S. Yang, Z. Pan, T. Amert, K. Wang, L. Yu, T. Berg, and M. Lin.
\newblock {Physics-Inspired Garment Recovery from a Single-View Image}.
\newblock \emph{ACM Transactions on Graphics}, 37\penalty0 (5):\penalty0 1--14,
  2018.

\bibitem[Zeng et~al.(2023)Zeng, Zeng, Tang, Wang, Yan, and Wang]{Zeng23}
X. Zeng, P. Zeng, C. Tang, P. Wang, B. Yan, and Y. Wang.
\newblock {Dbtrans: A Dual-Branch Vision Transformer for Multi-Modal Brain
  Tumor Segmentation}.
\newblock In \emph{Conference on Medical Image Computing and Computer Assisted
  Intervention}, pages 502--512, 2023.

\bibitem[Zhang et~al.(2023)Zhang, Liu, Ali, Zhao, Sun, Han, Liu, Zhai, Cui,
  Zhang, et~al.]{Zhang23e}
Xukun Zhang, Yang Liu, Sharib Ali, Xiao Zhao, Mingyang Sun, Minghao Han, Tao
  Liu, Peng Zhai, Zhiming Cui, Peixuan Zhang, et~al.
\newblock {Anatomical-Aware Point-Voxel Network for Couinaud Segmentation in
  Liver CT}.
\newblock In \emph{Conference on Medical Image Computing and Computer Assisted
  Intervention}, pages 465--474, 2023.

\bibitem[Zhuang(2018)]{Zhuang18a}
X. Zhuang.
\newblock {Multivariate Mixture Model for Myocardial Segmentation Combining
  Multi-Source Images}.
\newblock \emph{IEEE Transactions on Pattern Analysis and Machine
  Intelligence}, 41\penalty0 (12):\penalty0 2933--2946, 2018.

\end{thebibliography}

}

\clearpage

\appendix
\renewcommand{\thefigure}{S\arabic{figure}}
\renewcommand{\thetable}{S\arabic{table}}
\setcounter{figure}{0}
\setcounter{table}{0}

\setcounter{page}{1}
\maketitlesupplementary

\section{Geometric Regularization Loss Terms}
\label{sec:sup_reg_loss}

\noindent{\bf Edge Losses $ \mathcal{L}_{\text{Edge}}$ and $\mathcal{L}_{\text{EdgeUnif}}$. } We take them to be
\begin{align}
     \mathcal{L}_{\text{Edge}}^l & = \frac{1}{E} \sum_{i=1}^E |e_i| \; , \\
     \mathcal{L}_{\text{EdgeUnif}}^l & = \sqrt{ \frac{1}{E-1} \sum_{i=1}^E (|e_i| - \mathcal{L}^l_{\text{Edge}})^2 } \; ,  \nonumber
\end{align}
where $E$ is the total number of edges in the mesh and $e_i$ are its edges. Minimizing them  discourages the elongation of edges by penalizing both their average length and deviations from this average length, thereby promoting the regularity of the mesh facets. 

\noindent{\bf Normal Loss $\mathcal{L}_{\text{Norm}}$.} We write it as
\begin{align}
\mathcal{L}_{\text{Norm}} & = \sum_{i=1}^F \sum_{j=1}^F \text{Adj}(f_i, f_j) (1 - \cos (n_{f_j}, n_{f_i})) \; , \\
\text{Adj}(f_i, f_j) & =
    \begin{cases}
        1,\ \text{if}\ f_i, f_j\ \text{are adjacent faces} \\
        0,\ \text{otherwise}
    \end{cases} \; , \nonumber
\end{align}
where $F$ is the number of faces in the mesh and $f_i$, $f_j$ are faces. Minimizing it further regularizes pairs of adjacent triangle faces by harmonizing their normals.

\noindent{\bf Laplacian Loss $ \mathcal{L}_{\text{Lapl}}$.} We take it to be
\begin{align}
    \mathcal{L}_{\text{Lapl}} & = \frac{1}{V} \sum_{i=1}^V \Big\lVert v_i - \frac{1}{|N_i|} \sum_{j \in N_i} v_j \Big\rVert \; , \nonumber
\end{align}
where $v_i$ are the vertices, $V$ the number of vertices in the mesh and $N_i$ the set of indices of the vertices neighboring $v_i$. Minimizing it promotes smoothness by pushing each vertex towards the mean of its neighbors~\cite{Desbrun99}.

\section{Architecture and Preprocessing}
\label{sec:sup_impl}

For our segmentation network, we use nnU-Net in our experiments on the hippocampus and lung, with 5 levels containing [32, 64, 128, 256, 320] channels for the lung and 4 levels for the hippocampus. For our heart experiments, we use a smaller UNet with fewer channels which we find faster to train for comparable results. For the GCN, we use the same structure as in~\cite{Wickramasinghe20} with a hidden dimension of 32 and as many layers as the segmentation network decoders, and using learned neighborhood sampling~\cite{Wickramasinghe20}. We use the Adam optimizer with a learning rate of $10^{-4}$, $\beta_1 = 0.9$ and $\beta_2 = 0.999$. We set loss multipliers on a per-dataset basis by picking the best from a random search. All experiments are performed using a single V100 GPU.

For the MM-WHS-4 dataset, We preprocess each CT scan by cropping each heart around its bounding box with a 10\% extra margin added around the cube for padding. The resulting volume is resized to $128 \times 128 \times 128$ and values are normalized per instance to have zero-mean and unit standard deviation. 

For the MSD hippocampus we resize each volume to $64 \times 64 \times 64$ and normalize gray-level values using the mean and standard deviation of the whole training dataset. 

For the TS Lung dataset, we resize all volumes to $128 \times 128 \times 128$, and use a random 260/28/60 train/validation/test split for our experiments, with values being normalized per-instance. 

\section{Ablation Studies}
\label{sec:sup_ablation}

\subsection{Data-Based Template Generation}

We use the hand-designed templates of Figs.~\ref{fig:template1} and~\ref{fig:other-templates}, which we have to re-define for each new organ. Instead, we could derive them directly from labeled ground truth data. To demonstrate this, we triangulated  a representative lung segmentation  from MM-WHS-4, and decimated the result to generate approximately as many vertices as in our hand-designed template. We then trained our model to use this new template as we did before and evaluate on 4 out of 5 splits, to avoid evaluating on the split the segmentation comes from. In terms of reconstruction loss, using the data template underperforms only very slightly as shown in Tab.~\ref{tab:data-template-ablation}, but the reconstructed meshes are significantly less regular as shown in Fig.~\ref{fig:data-template-ablation}.

\newcommand{\meshzoom}[1]{%
    \begin{tikzpicture}
        \node[anchor=south west, inner sep=0] (main) at (0,0) {\includegraphics[width=4cm]{fig/#1}};

        \coordinate (A) at (1.5,1); %
        \coordinate (B) at (2.5,2); %
        \coordinate (C) at (2.5,1); %

        \draw[black, thick] (A) rectangle (B);

        \begin{scope}[xshift=5cm, yshift=0.5cm]
            \node[anchor=south west, inner sep=0] (zoom) at (0,0) {
                \includegraphics[width=2.5cm, clip, trim=750 500 750 676]{fig/#1_hd.png}
            };
            \draw[thick] (0,0) rectangle (2.5,2.5);
        \end{scope}

        \draw[-, black, thick] (C) -- (5.0,0.5);
        \draw[-, black, thick] (B) -- (5.0,3.0);
    \end{tikzpicture}
}

\begin{figure}[th]
    \centering
    \subfloat[\label{fig:sub:data-sphere}]{
        \meshzoom{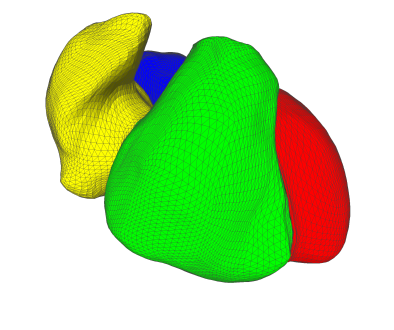}
    }\\
    \subfloat[\label{fig:sub:data-data}]{
        \meshzoom{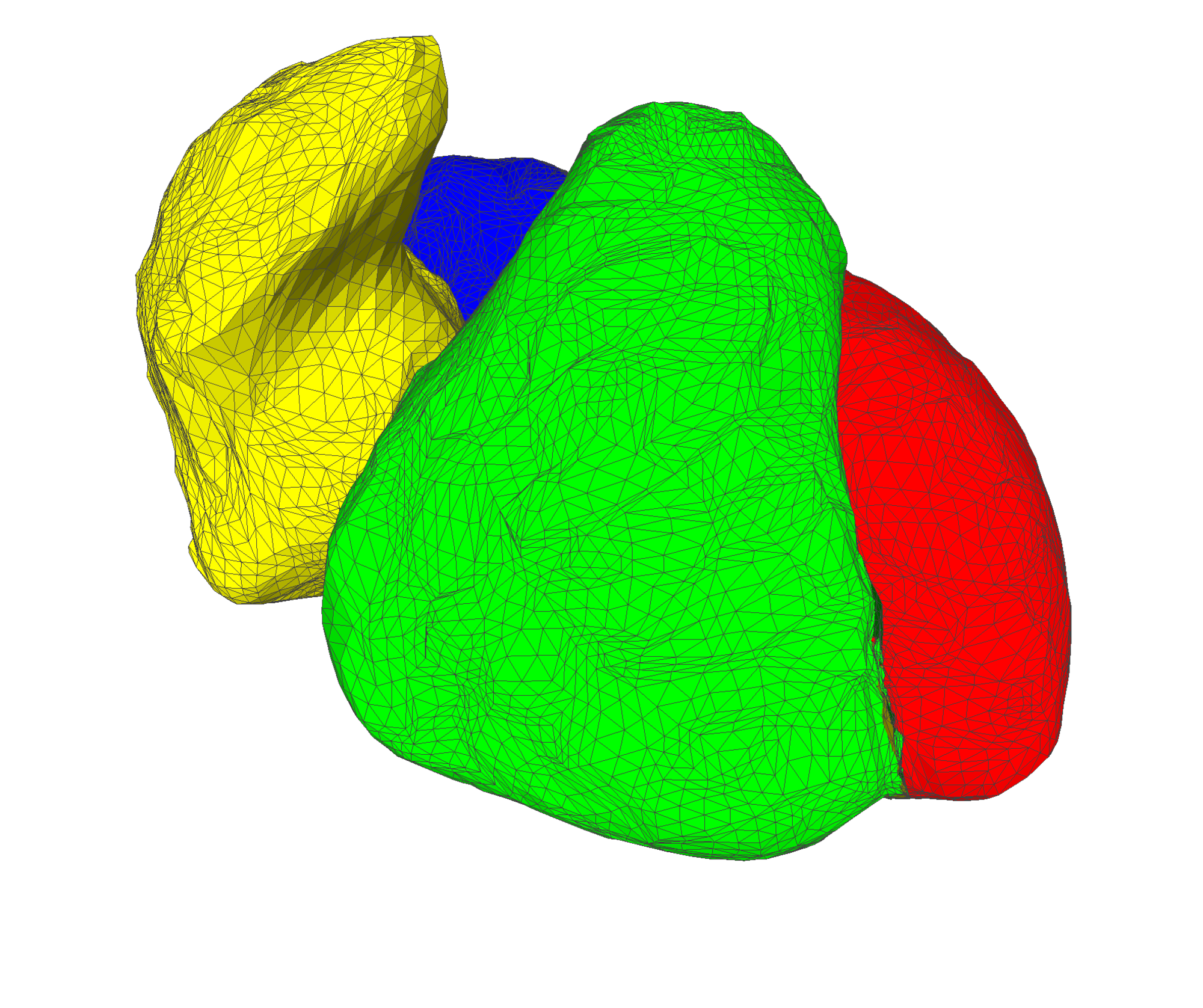}
    }
\vspace{-1mm}
\caption{\protect\textbf{Regular vs Data-Based Template.} (a) Reconstruction using the hand-crafted template of Fig.~\ref{fig:other-templates}(b). (b) Reconstruction using a data-based template. While it yields similar numerical reconstruction performance is similar, the triangles in the data-based version are much less regular and the resulting meshes less suitable for downstream applications, such as simulation.}
    \label{fig:data-template-ablation}
\end{figure}

\newcolumntype{?}{!{\vrule width 1pt}}

\begingroup
\setlength{\tabcolsep}{3pt}
\begin{table}
    \caption{regular template compared to a data template on MMWHS-4. \textit{Results are similar in terms of CD ($\times 10^{-3}$) despite visual differences.}} 
\vspace{-3mm}
\label{tab:data-template-ablation}
\centering
\renewcommand{\arraystretch}{1.2}
\resizebox{0.5\textwidth}{!}{
\begin{tabular}{@{}lccccc@{}}
\hline
 & \multicolumn{4}{c}{\textbf{Intersection Classes}} & \textbf{Other Classes} \\
\cmidrule(r){2-5} \cmidrule(l){6-6}
\textbf{Template} & LV $\cap$ LA & LV $\cap$ RV & RV $\cap$ RA & Average & Average \\
\cmidrule(r){1-1} \cmidrule(r){2-5} \cmidrule(l){6-6}
Sphere-Based & $0.68 \pm 0.44$ & $1.85 \pm 1.12$ & $\mathbf{2.30 \pm 1.24}$ & $\mathbf{1.61 \pm 0.68}$ & $\mathbf{1.18 \pm 0.30}$ \\
Data-Based    & $\mathbf{0.65 \pm 0.30}$ & $\mathbf{1.65 \pm 0.77}$ & $2.60 \pm 1.30$ & $1.63 \pm 0.55$ & $1.24 \pm 0.22$ \\
\hline
\end{tabular}
}

\end{table}
\endgroup

\subsection{Shared-Surface Supervision}

While we advocate training by separately supervising the shared surfaces, we could instead use a simpler setup where we keep the unified template, but supervise only the main classes. In this case, the shared surface components are supervised additively by both main classes they belong to. We report the resulting reconstruction performance on a single split of MM-WHS-4 in Tab.~\ref{tab:precise-supervision-ablation}.  There is a significant drop in the performance of the shared-surfaces. This degradation arises from the ambiguity and loss of identity of the shared surfaces during training. As they are not explicitly supervised, parts of the main classes can "leak" to cover parts of the ground truth shared surface, leaving the shared surface partially reconstructed in the deformed template.

\newcolumntype{?}{!{\vrule width 1pt}}

\begingroup
\setlength{\tabcolsep}{2pt}
\begin{table}
\caption{The effect of not supervising shared-surfaces in MMWHS-4. \textit{Performance is greatly reduced in terms of Chamfer distance ($\times 10^{-3}$)}}
\vspace{-3mm}
\label{tab:precise-supervision-ablation}
\centering
\renewcommand{\arraystretch}{1.2}
\resizebox{0.5\textwidth}{!}{
\begin{tabular}{@{}lccccc@{}}
\hline
 & \multicolumn{4}{c}{\textbf{Intersection Classes}} & \textbf{Other Classes} \\
\cmidrule(r){2-5} \cmidrule(l){6-6}
\textbf{Template} & LV $\cap$ LA & LV $\cap$ RV & RV $\cap$ RA & Average & Average \\
\cmidrule(r){1-1} \cmidrule(r){2-5} \cmidrule(l){6-6}
Shared-surface & 0.31 & 2.59 & 0.75 & 1.22 & 1.05 \\
Main-surface    & 0.57 & 22.07 & 2.06 & 8.23 & 1.09 \\
\hline
\end{tabular}
}

\end{table}
\endgroup

\end{document}